\documentclass[10p,twocolumn,twoside]{IEEEtran}
\pdfoutput=1

\usepackage{times}
\usepackage{graphicx}
\usepackage{amsmath}
\usepackage{amssymb}
\usepackage{booktabs}
\usepackage{balance}
\usepackage{multirow}
\usepackage{xspace}
\usepackage{verbatim}
\usepackage{enumerate}
\usepackage{algorithmic}
\usepackage[pagebackref=false,breaklinks=true,colorlinks,bookmarks=false,linkcolor=blue,citecolor=blue,pdftitle={On Robust Face Recognition via Sparse Encoding: the Good, the Bad, and the Ugly},pdfauthor={Yongkang Wong, Mehrtash T. Harandi, Conrad Sanderson}]{hyperref}

\usepackage{lineno}
\usepackage{color}

\graphicspath{{./}{./figures/}}

\def\Vec#1{{\boldsymbol{#1}}}
\def\Mat#1{{\boldsymbol{#1}}}

\newcommand{\ie}{{ie.}\@\xspace}
\newcommand{\eg}{{eg.}\@\xspace}
\newcommand{\etal}{{et~al.}\@\xspace}

\begin{document}

\pagestyle{empty}

\title
	{
	On Robust Face Recognition via Sparse Encoding:
  the Good, the Bad, and the Ugly
	}

\author
  {
  {Yongkang Wong,
  Mehrtash T. Harandi, 
  Conrad Sanderson}
  \thanks
    {
    \hrule
    
    \hspace{1ex}
    
    This paper is a revised and extended version of our earlier work~\cite{Yong_IJCNN_2012}.
    }
	
	~\\
  SeSaMe Centre, National University of Singapore, Singapore\\
  NICTA, GPO Box 2434, Brisbane, QLD 4001, Australia\\
  University of Queensland, School of ITEE, QLD 4072, Australia\\
  Queensland University of Technology, Brisbane, QLD 4000, Australia
  }
  
\maketitle
\thispagestyle{empty}

\begin{abstract}

In the field of face recognition,
Sparse Representation (SR) has received considerable attention during the past few years.
Most of the relevant literature focuses on holistic descriptors in closed-set identification applications.
The underlying assumption in SR-based methods is that each class in the gallery has sufficient samples and the query lies on the
subspace spanned by the gallery of the same class.
Unfortunately, 
such assumption is easily violated in the more challenging face verification scenario,
where an algorithm is required to determine if two faces (where one or both have not been seen before) belong to the same person.
In this paper, 
we first discuss why previous attempts with SR might not be applicable to verification problems.
We then propose an alternative approach to face verification via SR.
Specifically,
we propose to use explicit SR encoding on local image patches rather than the entire face.
The obtained sparse signals are pooled via averaging to form multiple region descriptors,
which are then concatenated to form an overall face descriptor.
Due to the deliberate loss spatial relations within each region (caused by averaging),
the resulting descriptor is robust to misalignment and various image deformations.
Within the proposed framework,
we evaluate several SR encoding techniques:
\mbox{$l_1$-minimisation}, 
Sparse Autoencoder Neural Network (SANN),
and an implicit probabilistic technique based on Gaussian Mixture Models.
Thorough experiments on AR, FERET, exYaleB, BANCA and ChokePoint datasets show that the proposed local SR approach obtains considerably better
and more robust performance than several previous state-of-the-art holistic SR methods,
in both verification and closed-set identification problems.
The experiments also show that 
\mbox{$l_1$-minimisation} based encoding has a considerably higher computational cost when compared to SANN-based and probabilistic encoding,
but leads to higher recognition rates.

\end{abstract}

\section{Introduction}
\label{sec:introduction}

Face based identity inference
(normally known by the all-encompassing term ``face recognition''),
can be generalised into three distinct configurations:
closed-set identification, open-set identification, and verification~\cite{Cardinaux_TSP_2006}.
The task of closed-set identification is to classify a given face as belonging to one of $K$ previously seen persons in a gallery.
In such configuration,
identification performance can be maximised by utilising class labels.
For example,
Linear Discriminant Analysis (LDA)~\cite{Belhumeur_PAMI_1997} separates the gallery such that small
within-class scatter and large between-class scatter are achieved.
However, 
this closed-set identification task assumes impostor attacks do not exist
and that each probe face must match a person in the gallery.
This is a necessarily limiting assumption
(as the gallery cannot cover all people in existance),
and hence algorithms specifically relying on the closed-set assumption do not readily translate to real-world applications~\cite{Doddington_SC_2000}.
In contrast,
both open-set identification and verification
explicitly take into account the possibilities of impostor attacks and previously unseen people.
In open-set identification,  a given face is assigned to one of $K+1$ classes,
with the extra class representing an ``unknown person''.
The task of verification is to determine if two given faces (or two face sets) belong to the same person,
where one or both identities may not have been observed beforehand.

Verification can be implemented as a pair-wise comparison, resulting in a distance or probability
that is then thresholded to achieve the final decision (which is a binary yes/no).
As such, open-set identification can be decomposed into a set of verification tasks
(one for each person in the gallery),
as long as the pair-wise verification distances or probabilities are used instead of the verification decisions.
In addition to the task of biometric user authentication~\cite{Cardinaux_TSP_2006,Doddington_SC_2000},
the ability to handle previously unseen people is useful in video surveillance~\cite{Ali_forensic_2010},
for applications such as person re-identification across multiple cameras~\cite{tu_reidentification_2007}.

Wright~\etal~\cite{Wright2009_PAMI} recently proposed Sparse Representation based Classification (SRC) 
for face identification problems.
The underlying idea is to represent a query sample {\small $\Vec{y}$} as a sparse linear combination of a dictionary {\small $\Mat{D}$},
where the dictionary usually contains holistic face descriptors.
Moreover, 
it is assumed that each subject has sufficient samples in the dictionary to span over possible subspaces.
Each probe image can be considered to be represented by a sparse code that is comprised of coefficients
that linearly reconstruct the image via the dictionary.
As such, it is expected that only those atoms in the dictionary that truly match the class query
sample contribute to the sparse code.
Wright~\etal~\cite{Wright2009_PAMI} exploited this by computing a class-specific similarity measure. 
More specifically, 
they computed the reconstruction error of a query image to class $i$ by considering only the sparse
codes associated with the atoms of the $i$-th class. 
The class that results in the minimum reconstruction error specifies the label of query.
To handle the case of a preson not present in the gallery,
the given query image is considered as an imposter
if the minimum reconstruction error exceeds a predefined threshold.
A more thorough discussion of class-based SRC can be found in~\cite{Yang_PR_2011}.

A significant body of literature was proposed with the aim of improving the original SRC. 
For example, 
Yang and Zhang~\cite{Yang_ECCV_2010} extended the original approach to use a holistic representation derived from Gabor features.
The Gabor-based SRC (GSRC) was shown to be relatively more robust against illumination changes as
well as small degree of pose mismatches. 
Another example is the Robust Sparse Coding (RSC) scheme proposed by Yang~\etal~\cite{Yang_CVPR_2011},
where sparse coding is modelled as a sparsity-constrained robust regression problem.
RSC was shown to outperform the original SRC and GSRC, 
as well as being more effective in handling of face occlusions. 
However,
RSC is computationally more expensive when compared to various SRC approaches.
Yang~\etal~\cite{Yang_ICCV_2011} explored the benefit of a structured dictionary, 
where each atom is associated to a class label.
Using the Fisher discrimination criterion~\cite{Fisher_AE_1936},
a set of class-specified sub-dictionaries is learned,
where each class has small within-class scatter and large between-class scatter.

In spite of the recent success in face identification,
SRC relies on the {\it sparsity} assumption.
The assumption holds when each class in the gallery has sufficient samples 
and the query lies on the subspace spanned by the gallery of the same class.
Shi~\etal~\cite{Shi_CVPR_2011} questioned the validity of the sparsity assumption for face data 
and showed that the assumption may be violated even in the identification scenario. 
Since in a verification system there might not be any mutual overlap between the probe faces and the
training data
(\ie~the probe identities were never seen by the system during training),
violation of the sparsity assumption is more likely to happen.
In other words, 
a verification system needs to be capable of making decisions even for classes it has not seen before.
This contradicts the sparsity assumption, 
and hence existing SRC approaches do not naturally extend to verification scenarios.

The majority of SR-based systems represent faces in a rigid and holistic manner~\cite{Wright2009_PAMI,Yang_PR_2011,Yang_ECCV_2010} 
(\ie~holistic descriptors).
That is,
each face is represented by one feature vector that describes the entire face
and implicitly embeds rigid spatial constraints between face components~\cite{Cardinaux_TSP_2006,Harandi_IJCV_2009}. 
Examples of such representation include
classic techniques such as PCA-based feature extraction~\cite{EIGENFACE_1991}.
Such treatment implies ideal image acquisition (\eg~perfect image alignment, perfect localisation/detection).
In reality, 
especially for fully automated systems,
attaining ideal images is very challenging (if not impossible) for low resolution moving
objects~\cite{low_res_face_detect01}. 
The adverse impact of imperfect face acquisition on recognition systems that utilise holistic face descriptors
has been demonstrated in~\cite{Cardinaux_TSP_2006,Rodriguez_IVC_2006}.

To tackle misalignment problems,
Wagner~\etal~\cite{Wagner_PAMI_11} recently proposed an SR-based face alignment algorithm.
Given a set of frontal training images and a query face image, 
{\small $\Vec{x}_{\operatorname{auto}}$}, 
extracted using an automatic face locator (detector), 
the algorithm finds the image transformation parameters which transform
{\small $\Vec{x}_{\operatorname{auto}}$} for the best reconstruction error.
Though this approach has shown promising results,
it can be criticised as being a computationally intensive method for correcting rigid face descriptors,
rather than tackling the source of the problem: rigid descriptors are inherently not robust to in-class face variations
(\eg~face expressions variations).

In contrast to rigid representations, 
a face can also be represented by a set of local features with relaxed spatial constraints\footnote
{
However,
it must be noted that not all local feature-based face representations automatically have relaxed spatial constraints. 
For example, in~\cite{Ekenel_EUSIPO_2005} local feature extraction is followed by concatenation
of the local feature vectors into one long vector.
The concatenation, in this case, effectively enforces rigid spatial constraints.
}. 
This allows for some movement and/or deformations of face components~\cite{Cardinaux_TSP_2006,Heisele_CVIU_2003,Sanderson_PR_2006}, 
and in turn leads to a degree of inherent robustness to expression and pose changes~\cite{Sanderson_PR_2006}, 
as well as robustness to misalignment
(where the misalignment is a byproduct of automatic face locators/detectors~\cite{Cardinaux_TSP_2006}).
Aharon~\etal~\cite{Aharon_TSP_2006} showed that local features satisfy the sparsity assumption when an overcomplete dictionary (trained from
a sufficient amount of samples) is presented.
Therefore,
in this paper we focus on the use of SR for encoding local features to handle the problem of imperfect image acquisition.

In the field of object recognition, 
{\it bag-of-words} (BoW) approaches~\cite{Csurka_ECCVW_2004,Lazebnik_CVPR_2006}
have been shown to be robust and effective for general image categorisation problems.
The underlying idea is to treat any given image as a set of local keypoints or patches,
followed by assigning each patch to a predetermined word with a vector quantisation (VQ) algorithm.
The given image can be represented as a vector of assignments, 
where each dimension of the vector indicates the count of patches assigned to a particular word.
In the field of face recognition, 
an extension of BoW for face images,
called Multi-Region Histograms (MRH),
represents each image as a concatenated set of regional probabilistic histograms~\cite{Sanderson_ICB_2009}.

We first note that VQ and probabilistic approaches to BoW representations
can be considered as a form of sparse coding~\cite{Coates_JMLR_2011}.
With this in mind,
we propose to employ more direct forms of SR within the MRH framework,
namely \mbox{$l_1$-minimisation} and a Sparse Autoencoder Neural Network (SANN).
We denote this approach as Locally Sparse Encoded Descriptor (LSED).
As shown later, 
LSED in conjunction with \mbox{$l_1$-minimisation} outperforms MRH
as well as previous holistic SR methods,
obtaining state-of-the-art performance in various identity inference configurations
(\ie both verification and identification).

\subsection{Contributions}
\label{sec:introduction_contribution}

There are four main contributions in this paper:
\begin{itemize}

\item
We briefly discuss why previous attempts with SR are not be applicable for verification tasks
and show a possible rudimentary extension of SR (with holistic face representation) to such tasks.

\item 
In contrast to following the traditional approach of using holistic face representation in conjunction with SR,
we explicitly use a local feature-based face representation (based on the well-established bag-of-words literature~\cite{Csurka_ECCVW_2004,Lazebnik_CVPR_2006,Sanderson_ICB_2009})
and employ SR to encode local image patches.
In addition to the probabilistic approach for SR implicitly used by MRH~\cite{Sanderson_ICB_2009},
we study the efficacy of two more direct SR techniques,
namely \mbox{$l_1$-minimisation} and Sparse Autoencoder Neural Network (SANN).

\item
Via thorough evaluations on face images captured in controlled and uncontrolled environment conditions,
as well as in various challenging situations such as pose mismatches, imperfect face alignment, blurring, etc.,
we show that the proposed local feature SR approach considerably outperforms state-of-the-art holistic SR approaches.
The experiments are conducted in both verification and closed-set identification setups.

\item
We analyse the computation cost of the proposed local feature SR approach in conjunction with various SR encoding techniques.
We show that $l_1$ encoding leads to the highest accuracy at the expense of considerably higher computation cost
than the second best technique, which is implicit SR encoding via probabilistic histograms.

\end{itemize}

We continue the paper as follows.
We first delineate the background theory of sparse encoding in Section~\ref{sec:background}.
In Section~\ref{sec:src_limitation}, we discuss how can holistic SR approaches be applied for face verification.
In Section~\ref{sec:algorithm}, we present and discuss the proposed LSED.
Section~\ref{sec:experiments_still} is devoted to experiments on various identity inference experiments using still images.
Image set matching experiments are given in Section~\ref{sec:experiments_image_set}.
Section~\ref{sec:conclusions} provides the main findings.
\section{Background Theory}
\label{sec:background}

In this section, 
we delineate the background theory of three sparse encoding approaches,
namely:
{\bf (a)}~\mbox{$l_1$-minimisation},
{\bf (b)}~Sparse Autoencoder Neural Network (SANN),
and
{\bf (c)}~probabilistic approach. 
Consider a finite training set 
{\small ${ \Mat{Y} = \left[ \; \Vec{y}_1, \; \Vec{y}_2, \; \cdots, \; \Vec{y}_M \; \right] \in \mathbb{R}^{d \times M} }$}.
Each sparse encoding approach requires a dictionary (or model),
{\small $\Mat{D} \in \mathbb{R}^{d \times N}$},
where each column {\small $\Vec{d}_i \in \mathbb{R}^{d}$} is called an atom.
Given the learned dictionary {\small $\Mat{D}$},
a probe vector {\small $\Vec{x}$} is then encoded as a sparse code {\small $\Vec{\widehat{\alpha}}$} by a chosen encoding scheme.

\subsection{Sparse Encoding via $l_1$-minimisation}
\label{sec:background_l1}

Given the trained overcomplete dictionary {\small $\Mat{D}$} and a probe vector {\small $\Vec{x} \in \mathbb{R}^{d}$} that is compressible,
a sparse solution {\small $\Vec{\widehat{\alpha}} \in \mathbb{R}^{N}$} exists such that {\small $\Vec{x}$} can be reconstructed with
small residual.
The sparse solution {\small $\Vec{\widehat{\alpha}}$} can be found by solving the following {\small $l_0$}-minimisation problem:%
\begin{equation}
  \min \| \Vec{\alpha} \|_0
  \;
  \operatorname{subject \; to}
  \;
  \| \Mat{D}\Vec{\alpha} - \Vec{x} \|_2^2 \; \leq \; \epsilon
\label{eqn:l0_sparse}
\end{equation}

\noindent
where the notation {\small $\| \Vec{\alpha} \|_0$} counts the nonzero entries of {\small $\Vec{\alpha}$} and
$\epsilon$ is the threshold for the reconstruction error \mbox{\small $\| \Mat{D}\Vec{\alpha} - \Vec{y} \|_2^2 \;$}. 

Solving the {\small $l_0$}-minimisation problem is NP-hard and difficult to approximate.
As shown in~\cite{Tropp_IEEE_2010},
the solution of Eqn.~(\ref{eqn:l0_sparse}) can be approximated with the following \mbox{$l_1$-minimisation} 
(aka convex relaxation) problem:%
\begin{equation}
  \min \| \Vec{\alpha} \|_1
  \;
  \operatorname{subject \; to}
  \;
  \| \Mat{D}\Vec{\alpha} - \Vec{x} \|_2^2 \; \leq \; \epsilon
\label{eqn:l1_sparse}
\end{equation}

\noindent
which can be solved in polynomial time by linear programming methods~\cite{Wright2009_PAMI,Chen_SIAM_2001}.
Another popular choice of sparse approximation technique is called the greedy pursuit approach,
which approximates the sparse solution through iterative local approximation.
However, 
the greedy pursuit approach can only produce the optimal solution under very strict conditions~\cite{Tropp_TIT_04},
whereas the convex relaxation has proven to be able to produce optimal or near optimal solutions for variety of
problems~\cite{Tropp_IEEE_2010}.

As discussed in~\cite{Coates_ICML_2011},
the choice of the dictionary learning algorithm has minor influence to the performance of a selected sparse encoding algorithm.
Therefore, 
the aforementioned \mbox{$l_1$-minimisation} problem can be coupled with any dictionary learning algorithm.
In this paper, 
we train the dictionary {\small $\Mat{D}$} using the K-SVD algorithm~\cite{Aharon_TSP_2006},
which is effective for representing small image patches for sparse encoding problems~\cite{Rubinstein_IEEE_2010}.
The algorithm first initialises a random dictionary {\small $\Mat{D}$} with {\small $l_2$} normalised atoms and performs an iterative two
stage process until convergence.
The objective function is to minimise the following cost function:
\begin{equation}  
  \underset{\Mat{D},\Mat{\alpha}}{\min} \| \Mat{Y} - \Mat{D} \Mat{\alpha}^{\operatorname{train}} \|_F^2
  \;
  \operatorname{subject \; to}
  \;
  \forall i, \; \| \Mat{\alpha}_i^{\operatorname{train}} \|_0 \; \leq \; T_0
\label{eqn:ksvd_algorithm}
\end{equation}

\noindent
where the notation{\small  $\| A \|_F$} stands for the Frobenius norm,
with {\small $\| A \|_F^2$} is defined as {\small $\sum\nolimits_{i} \sum\nolimits_{j} | a_{i,j} |^2$}.

The first stage (sparse coding stage), 
with dictionary {\small $\Mat{D}$},
the representation vectors {\small $\Vec{\alpha}_i^{\operatorname{train}}$} in Eqn.~(\ref{eqn:ksvd_algorithm})
are obtained using any pursuit algorithm~\cite{Tropp_SP_2006}.
In the second stage (dictionary update stage),
the algorithm updates each atom, {\small $\Vec{d}_i$},
by first computing the overall representation error matrix, {\small $\Mat{E}_i$}, using:
\begin{equation}
  \Mat{E}_i = \Mat{Y} - \sum\nolimits_{j \neq i} \Vec{d}_j \Vec{\alpha}_j^{\operatorname{train}}
\label{eqn:ksvd_error}
\end{equation}

By restricting to use a subset of {\small $\Mat{E}_i$}, 
which corresponds to the training vectors that use the atom {\small $\Vec{d}_i$}, 
we obtain {\small $\Mat{E}_i^R$}.
Let {\small $\Mat{U} \Delta \Mat{V}^{T}$} represent the singular value decomposition of~{\small $\Mat{E}_i^R$}.
The updated version of atom {\small $\Vec{d}_i$} is then obtained as the first column of~{\small $\Mat{U}$}.

\subsection{Sparse Encoding via Sparse Autoencoder Neural Network}
\label{sec:background_SANN}

An Artificial Neural Network (NN) is a non-linear statistical approach
to modelling complex relationships between input and output data~\cite{Bishop_1995}.
A generic configuration of a NN normally contains an input layer, a number or hidden layers, and an output layer.
Each layer is comprised of a number of `neurons' or `nodes',
which are basic computational units that take an input vector, 
an intercept term $b$ (or a bias unit), and compute an output via:
\begin{equation}
  h_{\Mat{W},b}(\Vec{x}) = f \left( \sum\nolimits_{i=1}^{N} \Vec{w}_i \Vec{x} + b \right)
\label{eqn:NN}
\end{equation}

\noindent
where {\small $\Vec{w}_i$} is the weight associated to neuron {\small $i$}
and {\small $f\left( \cdot \right)$} is an activation function which maps the output to a fixed range.

The SANN~\cite{Ranzato_NIPS_2007,Goodfellow_NIPS_2009} is a NN for efficient feature encoding where the aim is to learn a sparse and
compressed representation for a set of training data.
More specifically, 
SANN can reconstruct the training data with small reconstruction error using a small set of nodes in the hidden layer.
Under the framework of SANN, 
we employ unsupervised model training to learn a hidden layer that consists of $N$ nodes,
which is parameterised with a weight {\small $\Mat{W} \in \mathbb{R}^{d \times N}$}
and bias {\small $\Vec{b} \in \mathbb{R}^{N}$}.
The back-propagation algorithm~\cite{Bishop_1995} can be used for training by minimising the following cost
function~\cite{Coates_JMLR_2011}:
\begin{equation}
  J(\Mat{W},\Vec{b}) = J_{\operatorname{error}} + 
                       J_{\operatorname{weight}} + 
                       \beta  J_{\operatorname{sparsity}}
\label{eqn:SANN_cost}
\end{equation}

\noindent
where
\begin{small} 
\begin{eqnarray}
  J_{\operatorname{error}}    
  & \hspace{-0.5ex} \mbox{=} \hspace{-0.5ex} & \frac{1}{M} \sum\limits_{i=1}^{M} 
      \left( 
      \frac{1}{2} \| \Vec{\widehat{x}}_i - \Vec{x}_i \|^2  
      \right) \\ 
  J_{\operatorname{weight}}   
  & \hspace{-0.5ex} \mbox{=} \hspace{-0.5ex} & \frac{\lambda}{2} \| \Mat{W} \|^2 \\
  J_{\operatorname{sparsity}}
  & \hspace{-0.5ex} \mbox{=} \hspace{-0.5ex} & \sum\limits_{i=1}^{N} \operatorname{KL} \left( \rho \; \| \; \widehat{\rho}_i \right) \\
  & \hspace{-0.5ex} \mbox{=} \hspace{-0.5ex} & \sum\limits_{i=1}^{N} 
        \left[ 
        \rho \log \left( \frac{\rho}{\widehat{\rho}_i} \right) +
        (1-\rho) \log \left( \frac{1-\rho}{1-\widehat{\rho}_i} \right)
        \hspace{-0.5ex} \right]
\label{eqn:SANN_cost_2}
\end{eqnarray}
\end{small}

\noindent
The cost functions 
{\small $J_{\operatorname{error}}$}, {\small $J_{\operatorname{weight}}$}, and {\small $J_{\operatorname{sparsity}}$} are respectively the
{\it square reconstruction error} term, {\it weight decay} term and {\it sparsity penalty} term.

{\small $J_{\operatorname{error}}$} minimises the overall reconstruction error,
with {\small $\Vec{\widehat{x}}_i$} denoting the reconstructed version of {\small $\Vec{x}_i$}~\cite{Goodfellow_NIPS_2009}. 
The regularisation term {\small $J_{\operatorname{weight}}$} decreases the magnitude of the weights to prevent overfitting.
{\small $J_{\operatorname{sparsity}}$} constrains the network to achieve low ``activation'',
where {\small $\operatorname{KL} \left( \rho \; \| \; \widehat{\rho}_i \right)$} is the Kullback-Leibler divergence between
{\small $\rho$} and {\small $\widehat{\rho}_i$}.
The parameter {\small $\rho$} controls the degree of sparsity and {\small $\widehat{\rho}_i$} is the average activation of hidden node $i$.
The parameter {\small $\beta$} in Eqn.~(\ref{eqn:SANN_cost}) controls the contribution of {\small $J_{\operatorname{sparsity}}$}
(typically equal to 3).

Given the trained SANN and a probe vector {\small $\Vec{x}$},
the elements of the sparse code 
$\Vec{\widehat{\alpha}} = \left[ \widehat{\alpha}_{1}, \widehat{\alpha}_{2}, \cdots, \widehat{\alpha}_{N} \right]$ are calculated using:
\begin{equation}
  \widehat{\alpha}_{i} = \operatorname{sig}(\Vec{w}_i^T \Vec{x} + b_i)
\label{eqn:sparse_encoding_SANN}
\end{equation}

\noindent
where {\small $\Vec{w}_i$} and {\small $b_i$} are the $i$-th weight and bias respectively.
The logistic sigmoid function {\small $\operatorname{sig}(t) = 1 / (1 + \exp(-t))$} 
maps the output to the range of $[0,1]$.
In contrast to the $l_1$-minimisation approach described previously,
SANN has the advantage of avoiding the minimisation problem during the sparse encoding stage,
resulting in a lower computational cost.

\subsection{Implicit Sparse Encoding via Probabilistic Approach}
\label{sec:background_prob}

In the context of probabilistic modelling,
vectors are assumed to be independent and identically distributed
(this assumption is often incorrect but necessary to make the problem tractable~\cite{Reynolds_bio_2009}).
By assuming the vectors obey a Gaussian distribution,
all data can be modeled as a mixture of Gaussians or Gaussian Mixture Model (GMM).
GMM is a parametric probability density function represented as a weighted sum of Gaussian component
densities~\cite{Cardinaux_TSP_2006,Reynolds_bio_2009,Bishop_2006}.
Given a probe vector {\small $\Vec{x}$} and a trained model with relatively large number of Gaussians,
the normalised likelihood of {\small $\Vec{x}$} belonging to each Gaussian can be represented as a sparse
code {\small $\Vec{\widehat{\alpha}}$} with:

\noindent
\begin{small} 
\begin{equation}
  \Vec{\widehat{\alpha}}
  =
  \left[
       \frac
         { w_1 p \left( \Vec{x} | \Vec{\mu}_1, \Vec{\Sigma}_1 \right) }
         { \sum\limits_{n=1}^{N} w_n p \left( \Vec{x} | \Vec{\mu}_n, \Vec{\Sigma}_n \right) }, 
       ~\cdots,        
       ~\frac
         { w_N p \left( \Vec{x} | \Vec{\mu}_N, \Vec{\Sigma}_N \right) }
         { \sum\limits_{n=1}^{N} w_n p \left( \Vec{x} | \Vec{\mu}_n, \Vec{\Sigma}_n \right) }
 \right]
\label{eqn:sparse_encoding_prob}
\end{equation}
\end{small}

\noindent
where

\noindent
\begin{equation}
  p \left( \Vec{x} | \Vec{\mu}_n, \Vec{\Sigma}_n \right)
  =
  \frac
  {
  \operatorname{exp}
  \left[ 
    - \frac{1}{2} 
    \left( \Vec{x} - \Vec{\mu}_n \right)^{T} 
    \hspace{-0.5ex} \Vec{\Sigma}_n^{-1} 
    \hspace{-0.5ex} \left( \Vec{x} - \Vec{\mu}_n \right) 
  \right]
  }
  { \left( 2 \pi \right)^\frac{d}{2} | \Vec{\Sigma}_n |^\frac{1}{2} }
\label{eqn:sparse_encoding_prob_2}  
\end{equation}

\noindent
is a multi-variate Gaussian function~\cite{Bishop_2006,Duda_2001}.
The variables {\small $w_n$}, {\small $\Vec{\mu}_n$}, and {\small $\Vec{\Sigma}_n$} are, 
respectively, 
the weight, mean vector and diagonal covariance for Gaussian $n$.
The dictionary is trained by first initialising the mean vectors with a $k$-means clustering algorithm followed by
the Expectation-Maximisation algorithm~\cite{Duda_2001}.
We note that most of the entries in sparse code {\small $\Vec{\widehat{\alpha}}$}
are typically not exactly zero but are small enough to be treated as zero.
\section{SR: Identification vs. Verification}
\label{sec:src_limitation}

In this section,
we first briefly review the SR-based classification methodology for face identification problems.
We then discuss why such methodology is not suitable for face verification problems
and delineate a rudimentary extension to allow the use of SR with holistic descriptors in such problems.
This rudimentary holistic approach is separate and distinct from using SR at the level of local patches.

\subsection{Holistic SR for Face Identification}
\label{sec:src_limitation_identification}

Consider a closed-set face identification problem with a gallery comprised of {\small $N$} samples.
Let \mbox{\small $\Mat{D} \in \mathbb{R}^{d \times N}$} be the dictionary comprising all samples in the gallery.
Given a query {\small $\Vec{x} \in \mathbb{R}^{d}$},
the sparse solution {\small $\Vec{\widehat{\alpha}}$} can be estimated by solving Eqn.~(\ref{eqn:l1_sparse}).
Using only the coefficients associated with the $i$-th class,
Wright~\etal~\cite{Wright2009_PAMI} computed the residual, {\small $r_i(\Vec{x})$}, using:

\begin{equation}
  r_i(\Vec{x})
  =
  \| \Vec{x} - \Mat{D} \delta_i(\Vec{\widehat{\alpha}}) \|_2^2
\label{eqn:src_residual}
\end{equation}%

\noindent
where {\small $\delta_i$} is a binary vector with the non-zero entries being associated to class $i$.
The identity of query {\small $\Vec{x}$} is assigned using the rule:
{\small ${\operatorname{identity}} (\Vec{x}) = {\operatorname{arg\,min}}_{i} \; r_i(\Vec{x})$}.
This classification methodology is also used in the Gabor-based SRC~\cite{Yang_ECCV_2010}
and RSC~\cite{Yang_CVPR_2011}.

\subsection{Rudimentary Extension of Holistic SR to Face Verification}
\label{sec:src_limitation_verification}

In the context of face verification,
the identities of probe faces may not be present in the gallery.
As such, 
the sparsity assumption is likely to be violated,
making the classification methodology described above not applicable to verification problems.

An alternative way to incorporate SR in verification problems is to use the sparse code
(\ie~{\small $\Vec{\widehat{\alpha}}$}) as a face descriptor.
Given a dictionary
{\small $\Mat{D}$} and two faces {\small $\Vec{x}_a,\Vec{x}_b \in \mathbb{R}^{d}$},
we first generate their respective sparse solutions {\small $\Vec{\widehat{\alpha}}_a$} and {\small $\Vec{\widehat{\alpha}}_b$} using
Eqn.~(\ref{eqn:l1_sparse}).
The similarity score between these descriptors can be calculated using:
\begin{equation}
  s_{\operatorname{SR}} ( \Vec{x}_a, \Vec{x}_b | \Mat{D} )
  =
  \operatorname{dist} \left( \Vec{\widehat{\alpha}}_a - \Vec{\widehat{\alpha}}_b \right)
\label{eqn:sr_verification}
\end{equation}

\noindent
where {\small $\operatorname{dist}( \cdot )$} is the distance function of choice,
such as Euclidean or Hamming distance,
with a smaller value indicating a higher similarity between {\small $\Vec{x}_a$} and {\small $\Vec{x}_b$}. 
The classification decision (\ie~whether {\small $\Vec{x}_a$} and {\small $\Vec{x}_b$} represent the same person)
can be obtained by comparing {\small $s_{\operatorname{SR}}$} to a decision threshold.

In the above approach, the sparse solutions can be obtained from holistic face representations,
such as PCA-based feature extraction~\cite{EIGENFACE_1991}.
We therefore denote this approach as {\it holistic SR descriptor}.
\section{Locally Sparse Encoded Descriptor}
\label{sec:algorithm}

In the previous section,
we have shown an extension of holistic SR to verification problems.
However, as shown later, the holistic SR descriptor delivers poor performance.
In this section, 
we present an alternative way to utilise sparse coding in verification problems.
Motivated by the benefits of local feature-based face representation and BoW approaches,
we introduce a face descriptor termed Locally Sparse Encoded Descriptor (LSED),
which can be seen as an extension of MRH~\cite{Sanderson_ICB_2009}.
In addition to the implicit probabilistic encoding used in the original MRH formulation,
we propose to use two more direct sparse encoding techniques:
\mbox{$l_1$-minimisation} and SANN,
described in Sections~\ref{sec:background_l1} and~\ref{sec:background_SANN}.
We continue this section by first describing the face encoding framework,
followed by brief discussions on the characteristics of each sparse encoding technique.
We then elaborate on how the descriptor can be used for discriminating faces.

\subsection{Framework}
\label{sec:algorithm_framework}

A given face image is first split into $R$ fixed size regions,
where each region covers a relatively large portion of the face image.
For region $r$, a set of low-dimensional feature vectors,
{\small $\Mat{X}_r = \{ \Vec{x}_{r,1},\Vec{x}_{r,2},\hdots,\Vec{x}_{r,n} \}$},
is attained by dividing the region into smaller patches 
{\small $\Mat{p}_{r,1},\Mat{p}_{r,2},\hdots,\Mat{p}_{r,n}$}.
To account for varying contrast caused by illumination changes,
each patch is normalised to have zero mean and unit variance.

From each normalised patch {\small $\Mat{\widehat{p}}_{r,i}$},
a low dimensional texture descriptor,
{\small $\Vec{x}_{r,i}$},
is obtained via 2D DCT decomposition~\cite{Gonzalez_2007}.
Preliminary experiments suggest that patches of size {\small $8 \times 8$} pixels with 75\% overlap 
(\ie~adjacent patches are overlapped by either {\small $6 \times 8$} or {\small$8 \times 6$ pixels})
lead to good performance~\cite{Sanderson_ICB_2009}.
Moreover, 
we selected the 15 lowest frequency components of the DCT coefficients,
with the zeroth coefficient discarded (as it has no information due to the aforementioned normalisation step).
We note that it is also possible to use other texture descriptors, 
such as raw pixels, Gabor wavelets~\cite{Lee_PAMI_2005} and Local Binary Patterns~\cite{Ahonen_PAMI_2006}.
Preliminary experiments suggest that the DCT-based texture descriptors lead to better performance.

\begin{figure*}[!tb]
\centering
  \includegraphics[width=2.0\columnwidth]{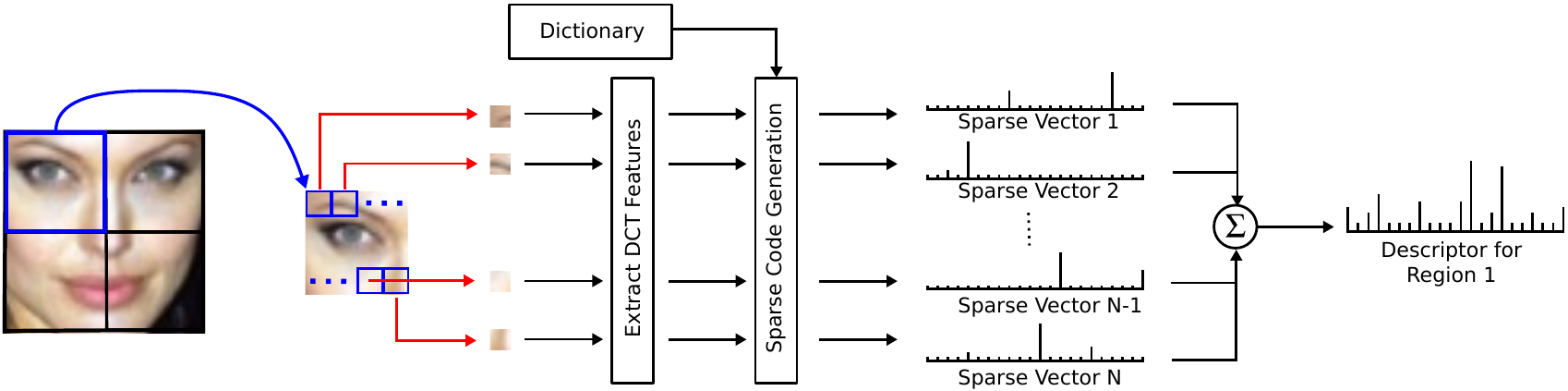}
  \caption
    {
    Conceptual demonstration of the LSED framework.
    A given face image is divided into regions,
    followed by breaking each region into smaller patches.
    For each patch, a sparse vector is obtained by a sparse encoder using a learned dictionary.
    Each regional face descriptor is computed by pooling the sparse vectors from the corresponding region.
    }
  \label{fig:mrh_concept}
  \vspace{8ex}
\end{figure*}

Each $i$-th texture descriptor from region $r$, 
{\small $\Vec{x}_{r,i}$}, 
is then described by a sparse code {\small $\Vec{\widehat{\alpha}}_{r,i}$}.
In the original formulation of MRH~\cite{Sanderson_ICB_2009},
the sparse code is implicitly generated using the probabilistic encoding approach elaborated in Eqn.~(\ref{eqn:sparse_encoding_prob}).
Having each patch represented by a sparse code, 
each region $r$ is then described via the following pooling strategy:%
\begin{equation}
  \Vec{h}_r 
  = 
  \frac{1}{N_{p}} \sum\nolimits_{i=1}^{N_{p}} \widehat{\Vec{\alpha}}_{r,i} 
\label{eqn:avg_pooling}
\end{equation}

\noindent
where {\small $\widehat{\Vec{\alpha}}_{r,i}$} is the $i$-th sparse vector in region $r$
and {\small $N_{p}$} is the number of patches in region $r$.
Due to the averaging operation,
in each region there is a loss of spatial relations between face parts. 
As such, 
each region is in effect described by an orderless collection of local descriptors.
A conceptual diagram of the framework is shown in Figure~\ref{fig:mrh_concept}.

We propose to use two other sparse encoding techniques to generate the sparse code {\small $\Vec{\widehat{\alpha}}_{r,i}$},
namely, 
\mbox{$l_1$-minimisation} (using Eqn.~(\ref{eqn:l1_sparse})) and SANN (using Eqn.~(\ref{eqn:sparse_encoding_SANN})).
For the \mbox{$l_1$-minimisation} based encoding,
the generated sparse codes may consist of negative coefficients,
which causes a problem with the averaging pooling strategy in Eqn.~(\ref{eqn:avg_pooling}).
To address this, 
the patch level sparse codes can be obtained with nonnegative encoding~\cite{Bruckstein_TIT_08} or by splitting the positive and negative
coefficients into two sparse codes followed by vector concatenation~\cite{Coates_ICML_2011}.
In preliminary experiments
we found that the most robust performance can be obtained by simply applying an absolute function to each patch level sparse code.

\begin{figure*}[!tb]
  \begin{center}
    \begin{minipage}{\textwidth}
    
      \begin{minipage}{\textwidth}
        \centering
        \begin{minipage}{0.02\textwidth}
          \centerline{\small\bf (a)}
        \end{minipage}
        \hfill
        \begin{minipage}{0.46\textwidth}
          \centerline{\includegraphics[width=\columnwidth]{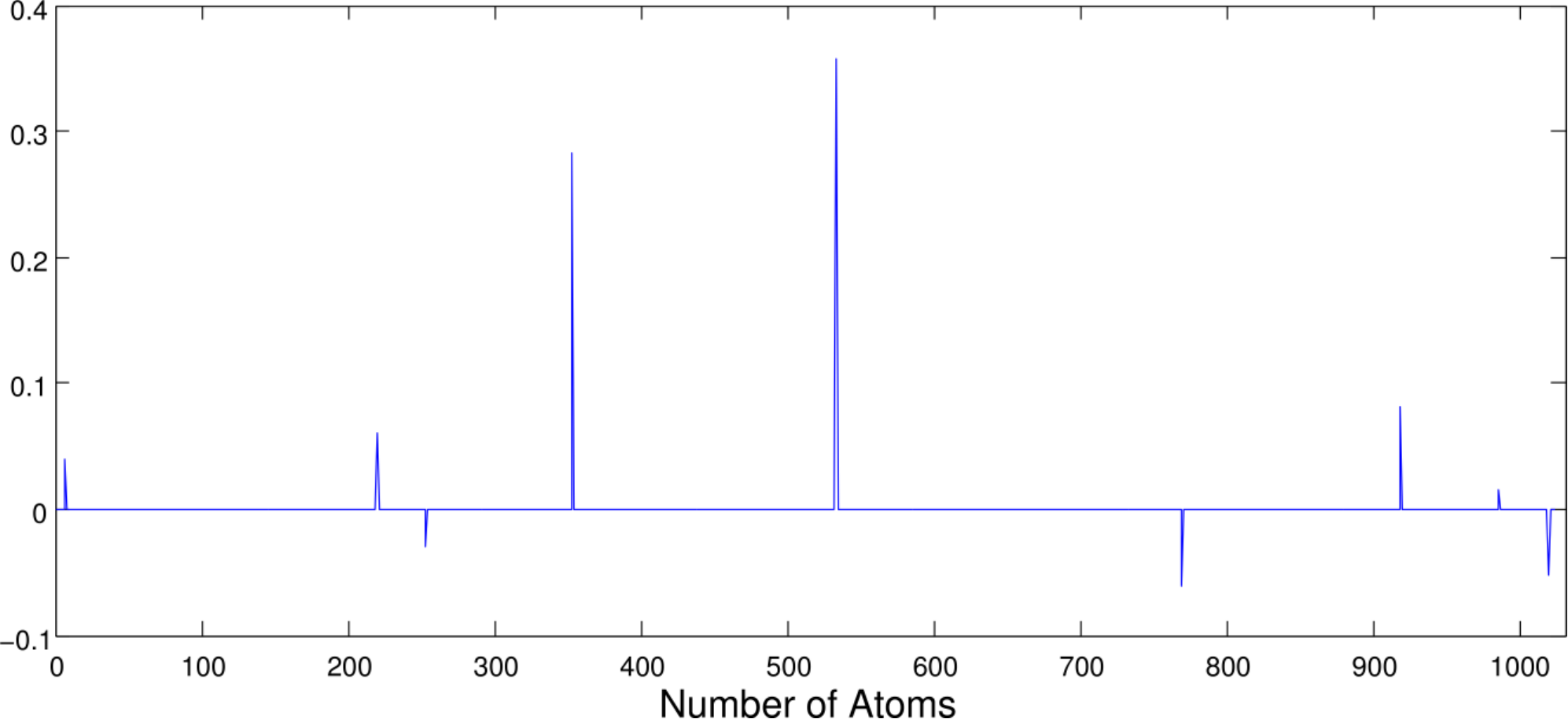}}
        \end{minipage}
        \hfill
        \begin{minipage}{0.46\textwidth}
          \centerline{\includegraphics[width=\columnwidth]{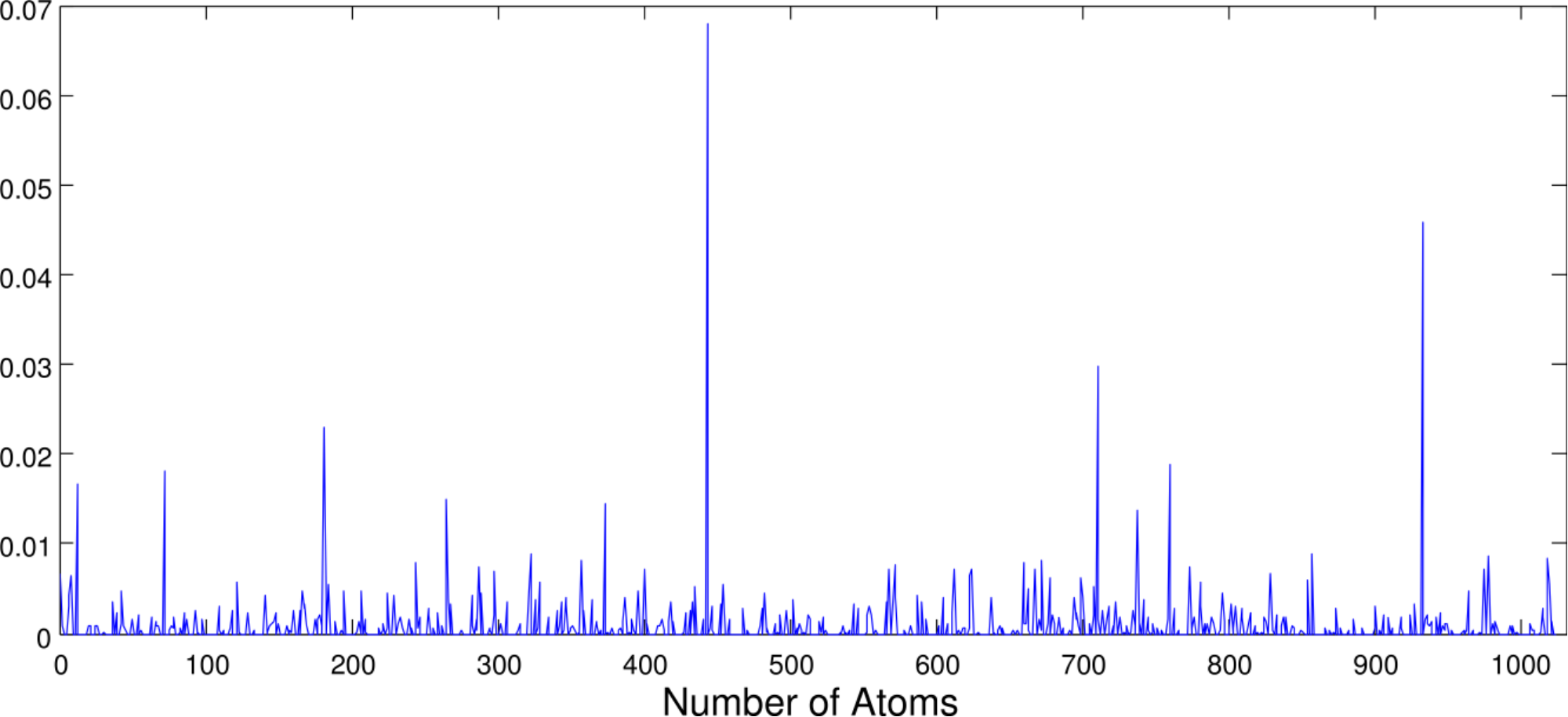}}
        \end{minipage}
      \end{minipage}

      \begin{minipage}{\textwidth}
        \centering
        \begin{minipage}{0.02\textwidth}
          \centerline{~}
          \centerline{~}
        \end{minipage}
        \hfill
        \begin{minipage}{0.46\textwidth}
          \centerline{~}
          \centerline{~}
        \end{minipage}
        \hfill
        \begin{minipage}{0.46\textwidth}
          \centerline{~}
          \centerline{~}
        \end{minipage}
      \end{minipage}
      
      \begin{minipage}{\textwidth}
        \centering
        \begin{minipage}{0.02\textwidth}
          \centerline{\small\bf (b)}
        \end{minipage}
        \hfill
        \begin{minipage}{0.46\textwidth}
          \centerline{\includegraphics[width=\textwidth]{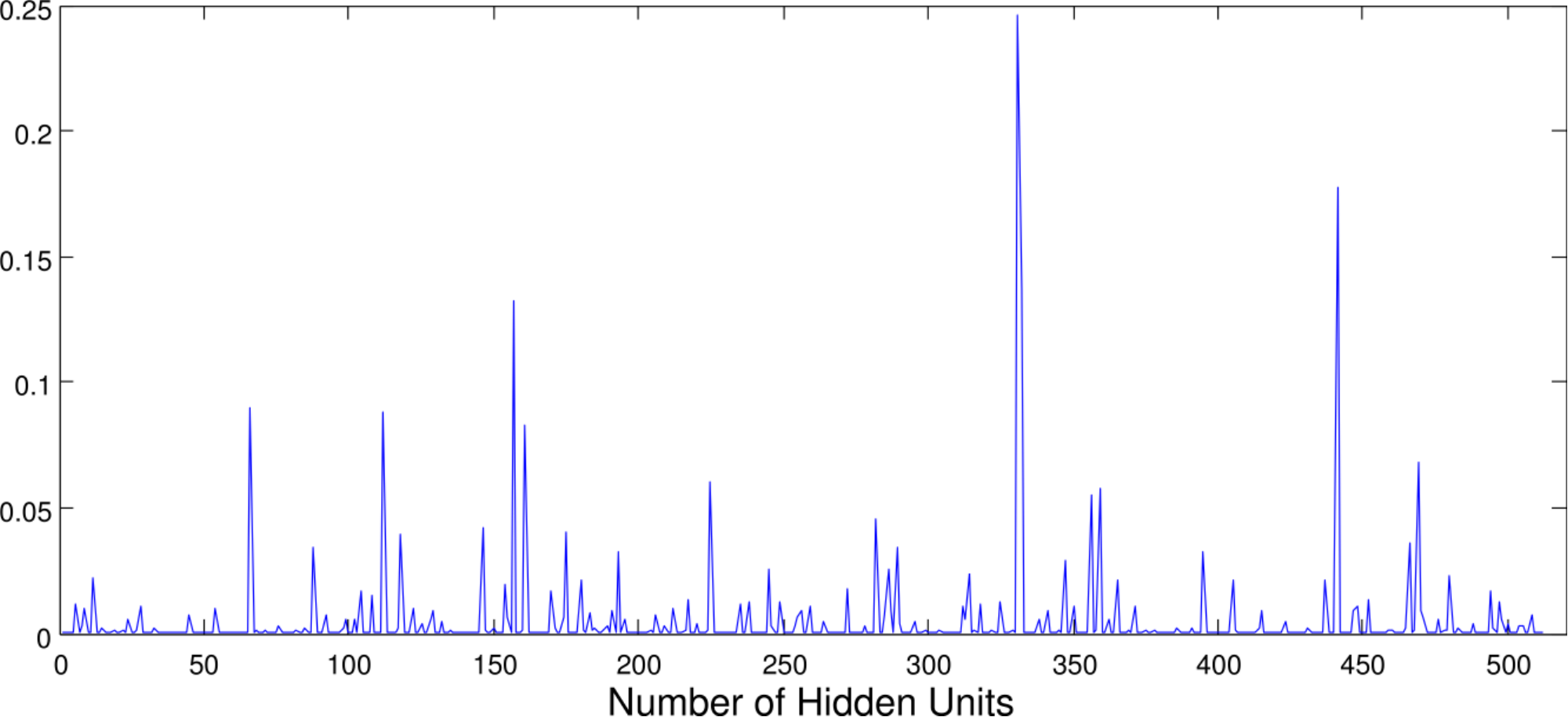}}
        \end{minipage}
        \hfill
        \begin{minipage}{0.46\textwidth}
          \centerline{\includegraphics[width=\columnwidth]{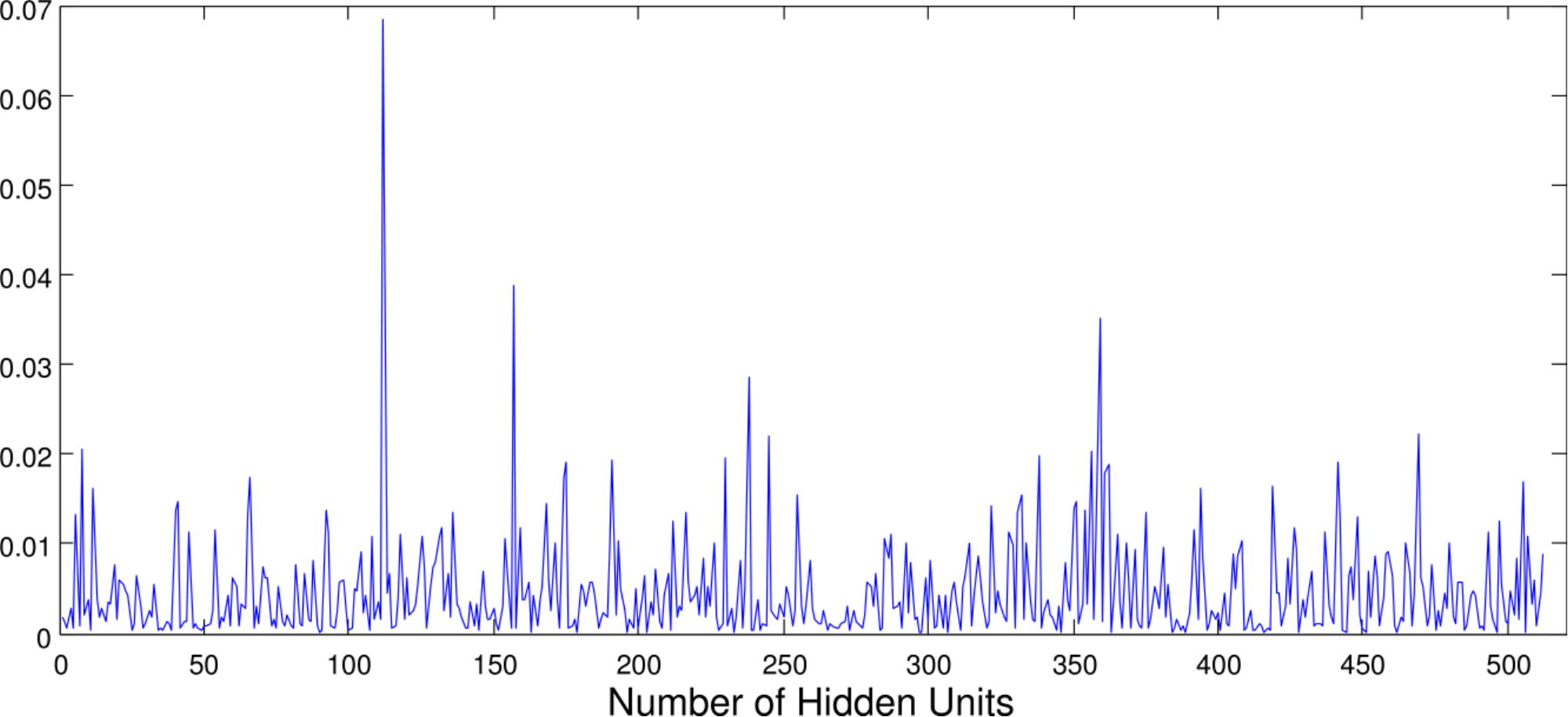}}
        \end{minipage}
      \end{minipage}
      
      \begin{minipage}{\textwidth}
        \centering
        \begin{minipage}{0.02\textwidth}
          \centerline{~}
          \centerline{~}
        \end{minipage}
        \hfill
        \begin{minipage}{0.46\textwidth}
          \centerline{~}
          \centerline{~}
        \end{minipage}
        \hfill
        \begin{minipage}{0.46\textwidth}
          \centerline{~}
          \centerline{~}
        \end{minipage}
      \end{minipage}

      \begin{minipage}{\textwidth}
        \centering
        \begin{minipage}{0.02\textwidth}
          \centerline{\small\bf (c)}
        \end{minipage}
        \hfill
        \begin{minipage}{0.46\textwidth}
          \centerline{\includegraphics[width=\columnwidth]{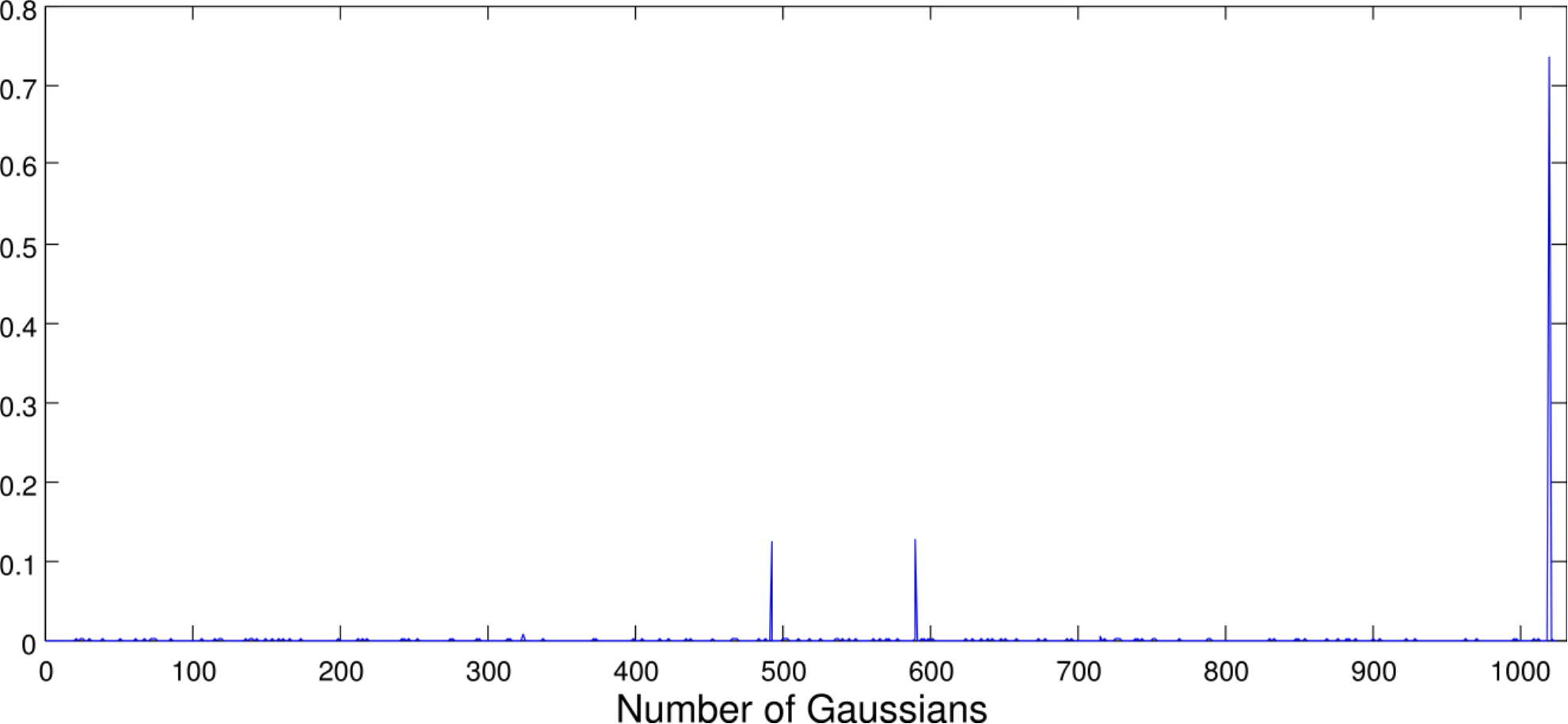}}
        \end{minipage}
        \hfill
        \begin{minipage}{0.46\textwidth}
          \centerline{\includegraphics[width=\columnwidth]{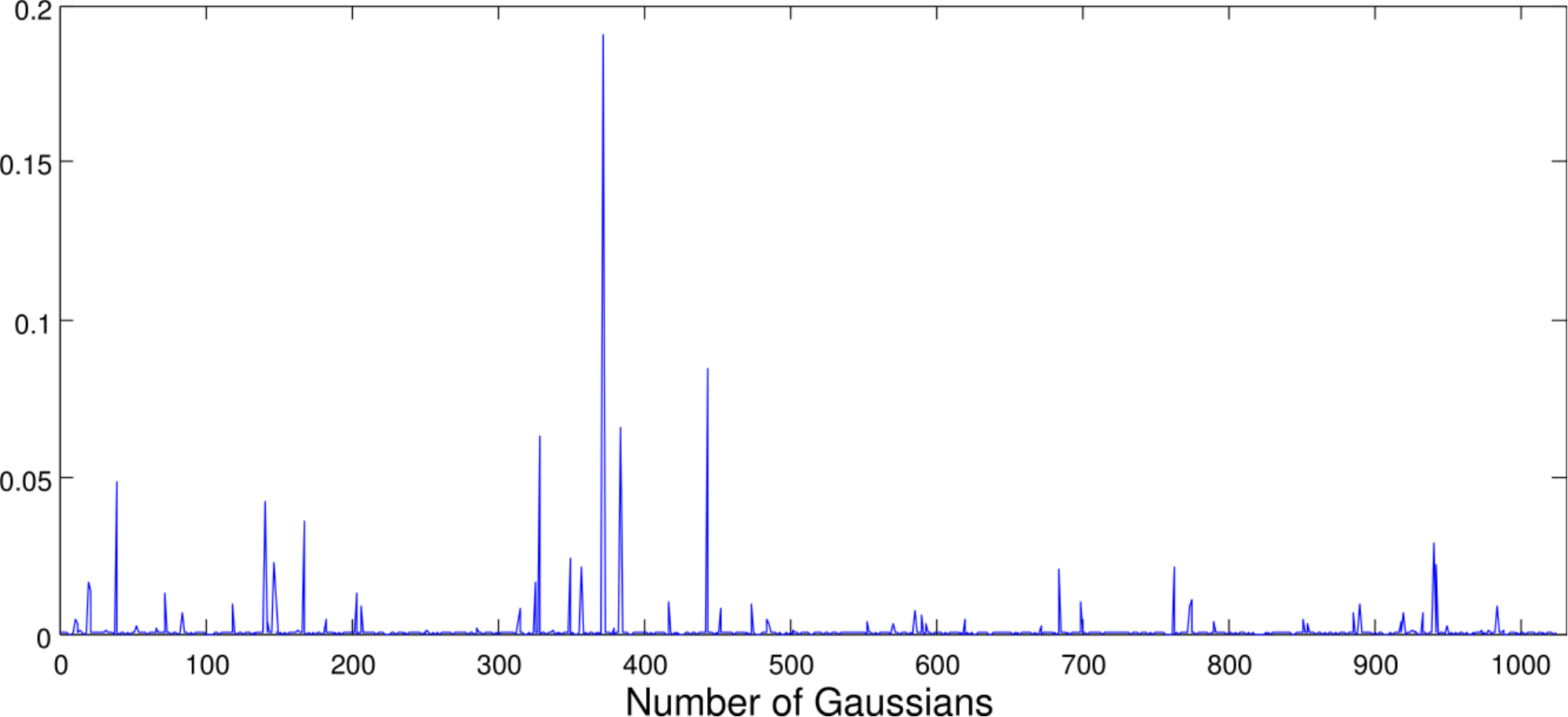}}
        \end{minipage}
      \end{minipage}

      \begin{minipage}{\textwidth}
        \centering
        \begin{minipage}{0.02\textwidth}
          \centerline{~}
          \centerline{~}
          \centerline{~}
        \end{minipage}
        \hfill
        \begin{minipage}{0.46\textwidth}
          \centerline{~}
          \centerline{~}
          \centerline{Patch level}
        \end{minipage}
        \hfill
        \begin{minipage}{0.46\textwidth}
          \centerline{~}
          \centerline{~}
          \centerline{Region level}
        \end{minipage}
      \end{minipage}

    \end{minipage}
  \end{center}
  \caption
    {
    Left column: examples of sparse codes for a single patch.
    Right column: examples of resultant region descriptors obtained via the pooling strategy in Eqn.~(\ref{eqn:avg_pooling}).
    Three sparse encoding approaches are shown:
    {\bf (a)}~\mbox{$l_1$-minimisation},
    {\bf (b)}~Sparse Autoencoder Neural Network,
    {\bf (c)}~probabilistic.
    For \mbox{$l_1$-minimisation} based encoding, 
    an absolute function is applied to each patch level code prior to applying the pooling strategy.
    For probabilistic encoding, most of the coefficients are not exactly zero but are small enough to be treated as zero.
    }
  \label{fig:LSED_example}
  \vspace{8ex}
\end{figure*}

The dictionary used by each sparse encoding approach is described in Section~\ref{sec:background}.
Examples of LSED with the three sparse encoding techniques are shown in Figure~\ref{fig:LSED_example},
where LSED with probabilistic encoding is the sparsest at both the patch level and the region level,
whereas the SANN-based encoding produces relatively noisier descriptors while maintaining a good degree of sparsity.
We discuss the differences of the encoding techniques below.

\subsection{Characteristics of Sparse Encoding}
\label{sec:algorithm_role}

In Section~\ref{sec:background},
we presented three sparse encoding approaches (\ie~\mbox{$l_1$-minimisation}, SANN and probabilistic encoding).
We note that there are some fundamental differences between the approaches.

The probabilistic approach computes the normalised likelihood using each Gaussian in the GMM,
which indirectly models each patch as a sparse vector.
The sparsity in this case stems from a very small subset of the Gaussians (typically 2 or~3) being close to a given sample.
The close Gaussians provide high normalised likelihoods, while the remaining Gaussians have likelihoods that are close to zero.

In contrast,
the \mbox{$l_1$-minimisation} approach solves an optimisation problem based on the reconstruction error 
(\ie~reconstruct a given patch as a linear combination of dictionary atoms),
with the optimal solution obtained for each patch.
The SANN-based approach uses a similar objective (\ie~patch reconstruction).
However, it avoids minimisation of the reconstruction error for each patch~\cite{Ranzato_NIPS_2007}.
The sparse solution for any given local patch is obtained by feeding the given patch into the SANN,
which is a very fast process that consists of straightforward linear algebra.
SANN assumes that the training samples provide the generic distribution of the data
and the optimisation is performed only on the training samples.
As such, this encoding approach may not deliver the optimal solution for any given patch.

\subsection{Similarity-Based Classification}
\label{sec:algorithm_classification}

Comparison between two faces is accomplished by comparing their corresponding regional descriptors.
Using the method from~\cite{Sanderson_ICB_2009},
the matching score between faces {\small $A$} and {\small $B$} can be calculated via:%
\begin{equation}
  s_\mathtt{raw} (A,B) 
  = 
  \frac{1}{R}
  \sum\nolimits_{r=1}^{R} 
  \left\| \Vec{h}_{r}^{[A]} - \Vec{h}_{r}^{[B]} \right\|_{1}
\label{eqn:raw_dist}
\end{equation}
 
\noindent
where {\small $R$} is the number of regions. 
To account for uncontrolled image conditions not already handled by the patch-based analysis,
a cohort normalisation~\cite{Doddington_SC_2000,Sanderson_ICB_2009} based distance can be employed:%
\begin{equation}
  s_\mathtt{norm} (A,B) =
  \frac
    {
    s_\mathtt{raw} (A,B)
    }
    {
    \sum\nolimits_{i=1}^{N_C} s_\mathtt{raw} (A, C_i) +
    \sum\nolimits_{i=1}^{N_C} s_\mathtt{raw} (B, C_i)
    }
\label{eqn:norm_dist}
\end{equation}

\noindent
where the cohort faces {\small $C_i$} are assumed to be reference faces
that are different from images of persons {\small $A$} or {\small $B$}.
To reach a decision as to whether faces {\small $A$} and {\small $B$} belong to the same person, 
{\small $s_\mathtt{norm} (A,B)$} can be compared to a decision threshold.

\begin{figure*}[!tb]
  \begin{minipage}{0.8\textwidth}
    \centering
    \begin{minipage}{1.0\textwidth}
      \begin{minipage}{1.0\textwidth}
        \begin{center}
          \parbox[b][16ex][c]{3ex}{\small\bf (a)}
          \includegraphics[width=0.23\textwidth]{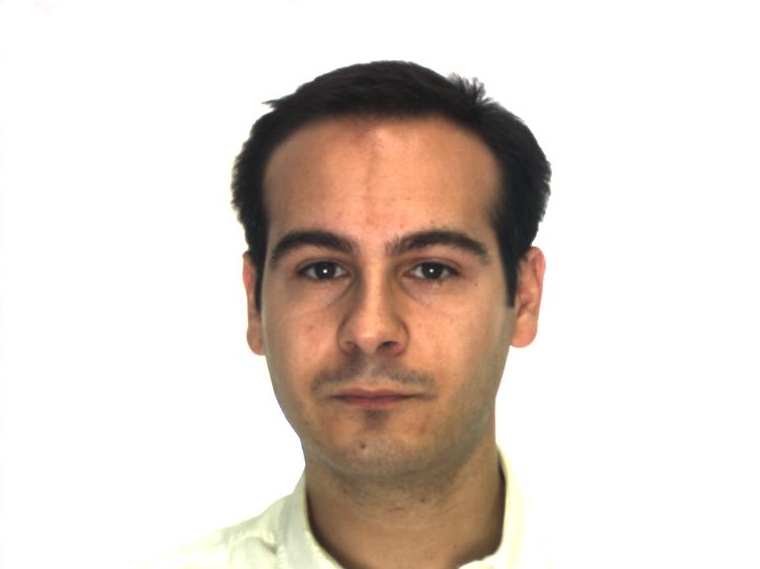}
          \includegraphics[width=0.23\textwidth]{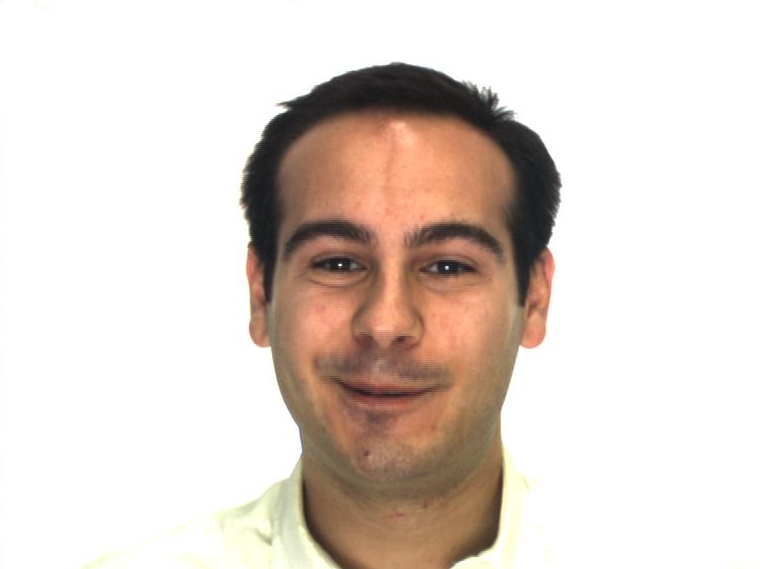}
          \includegraphics[width=0.23\textwidth]{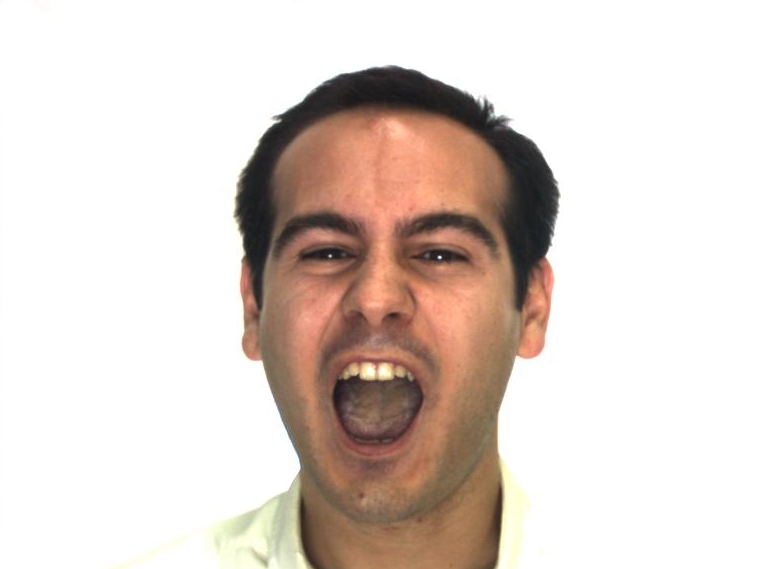}
          \includegraphics[width=0.23\textwidth]{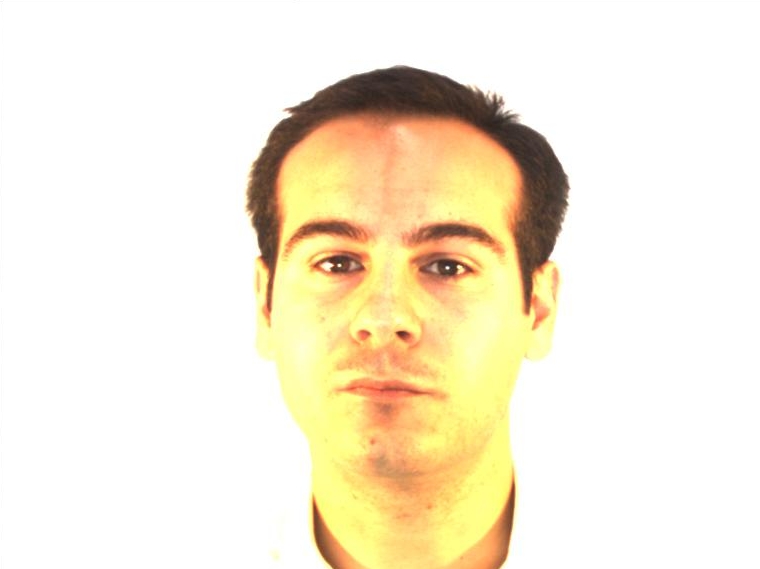}
        \end{center}
      \end{minipage}
      \begin{minipage}{1.0\textwidth}
        \centerline{\vspace{2px}}
      \end{minipage}
      \begin{minipage}{1.0\textwidth}
        \begin{center}
          \parbox[b][19ex][c]{3ex}{\small\bf (b)}
          \includegraphics[width=0.23\textwidth]{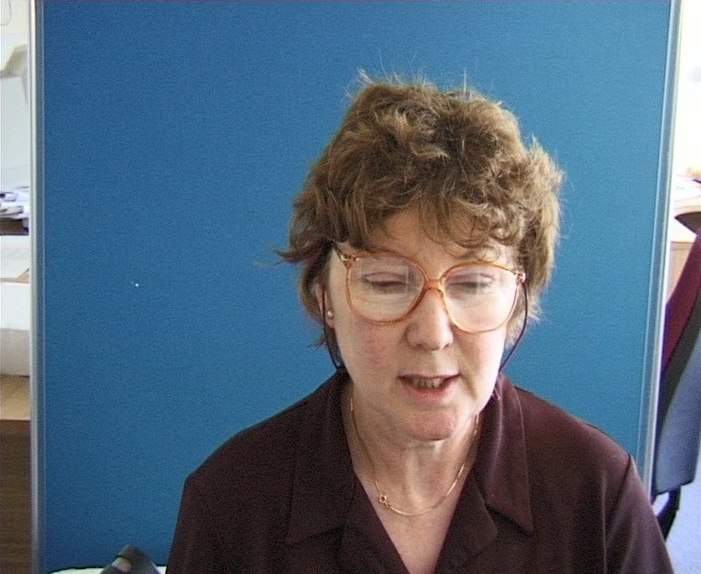}
          \includegraphics[width=0.23\textwidth]{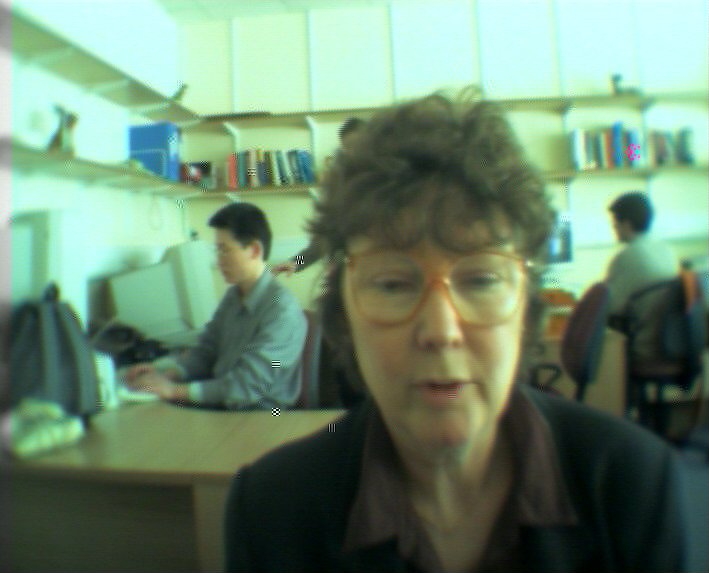}
          \includegraphics[width=0.23\textwidth]{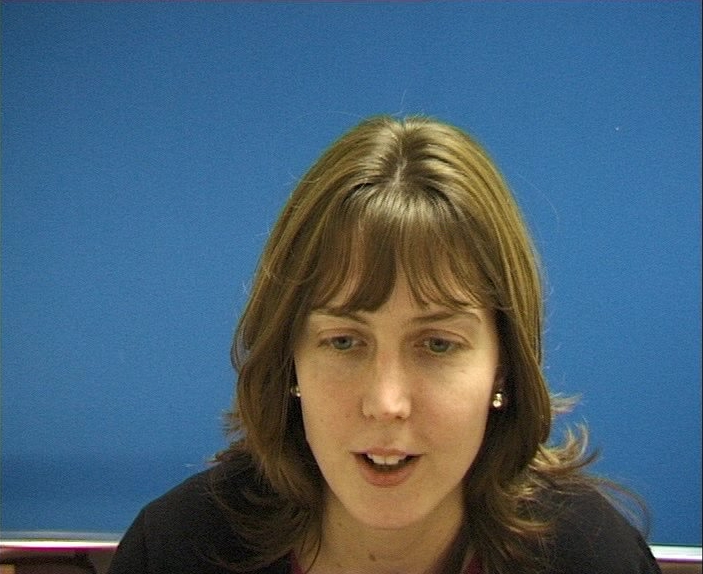}
          \includegraphics[width=0.23\textwidth]{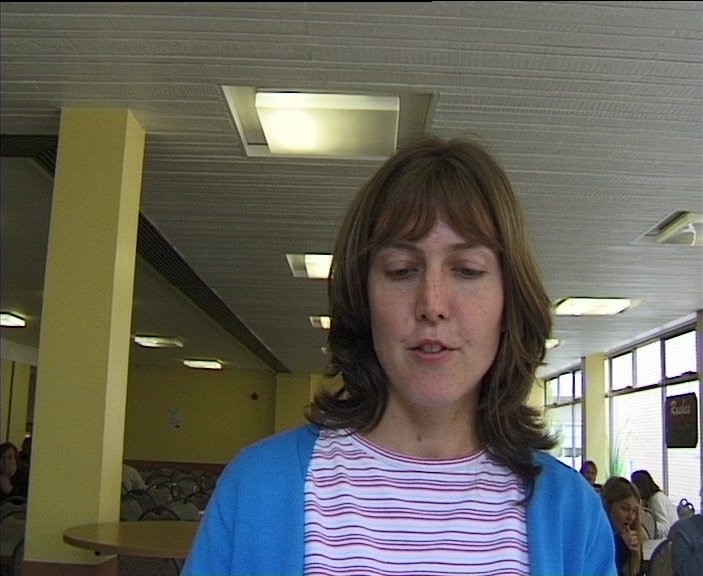}
        \end{center}
      \end{minipage}  
      \begin{minipage}{1.0\textwidth}
        \centerline{\vspace{2px}}
      \end{minipage}
      \begin{minipage}{1.0\textwidth}
        \begin{center}
          \parbox[b][14ex][c]{3ex}{\small\bf (c)}
          \includegraphics[width=0.23\textwidth]{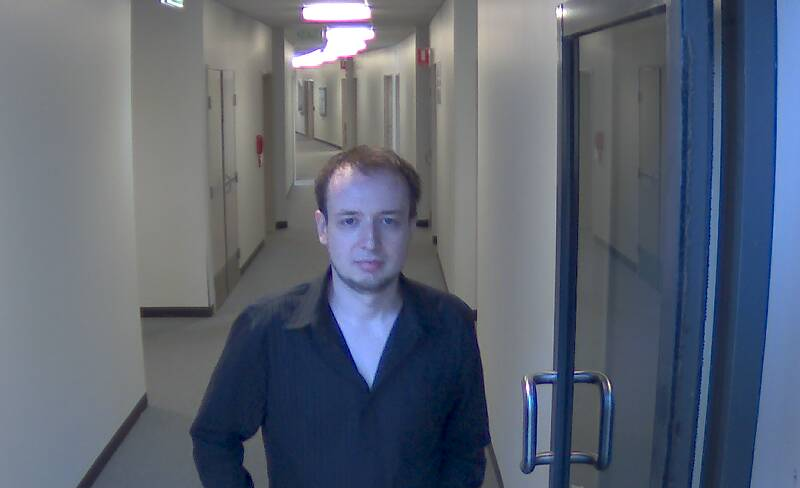}
          \includegraphics[width=0.23\textwidth]{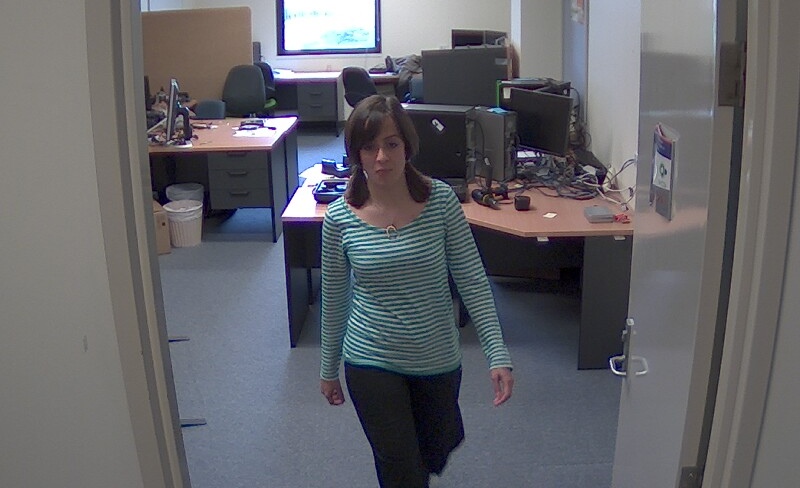}
          \includegraphics[width=0.23\textwidth]{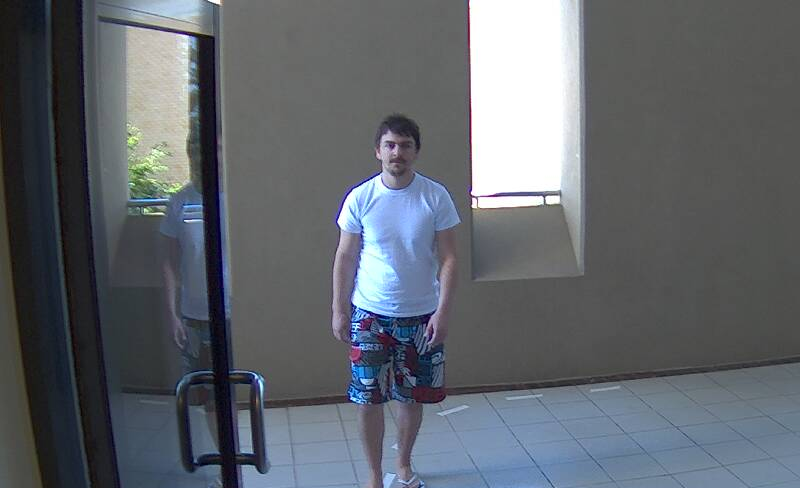}
          \includegraphics[width=0.23\textwidth]{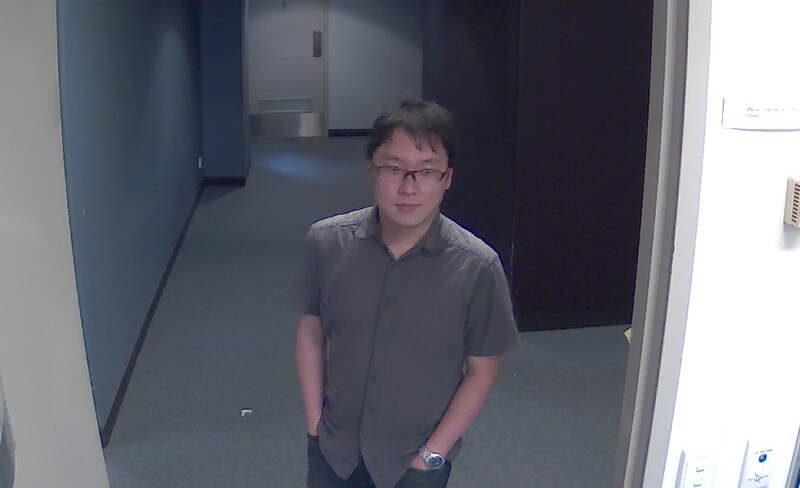}
        \end{center}
      \end{minipage} 
    \end{minipage}
    \caption
      {
      Example raw images from several datasets.
      {\bf (a)}~The AR dataset contains 14 images per subject with various expressions and lighting conditions.
      {\bf (b)}~The BANCA dataset: each subject was recorded under 3 scenarios: 
            {\it controlled} (columns 1 \& 3), 
            {\it degraded} (column 2), and 
            {\it adverse} (column 4).
      {\bf (c)}~The ChokePoint dataset contains 29 subjects captured in 4 distinct portals. 
      }
    \label{fig:face_images}
  \end{minipage}
  ~\vline~
  \begin{minipage}{0.18\textwidth}
    \centering
    \begin{minipage}{1\textwidth}
      \centerline{\includegraphics[width=0.45\textwidth]{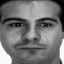}~\includegraphics[width=0.45\textwidth]{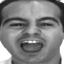}}
      \centerline{AR}
    \end{minipage}

    \centerline{~}
    \centerline{~}
    \centerline{~}
    \centerline{~}
      
    \begin{minipage}{1\textwidth}
      \centerline{\includegraphics[width=0.45\textwidth]{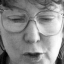}~\includegraphics[width=0.45\textwidth]{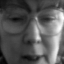}}
      \centerline{BANCA}
    \end{minipage}
    
    \centerline{~}
    \centerline{~}
    \centerline{~}
    
    \begin{minipage}{1\textwidth}
      \centerline{\includegraphics[width=0.45\textwidth]{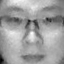}~\includegraphics[width=0.45\textwidth]{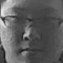}}
      \centerline{ChokePoint}
    \end{minipage}
    
    \centerline{~}
    
    \caption
      {
      Examples of cropped images.
      }
    \label{fig:face_crop_example}
    
  \end{minipage}
\end{figure*}

\section{Experiments with Still Images}
\label{sec:experiments_still}

In this section,
we examine the performance of LSED on several identity inference configurations:
{\bf (a)}~verification with various face alignment errors and sharpness variations,
{\bf (b)}~verification with pose mismatches,
{\bf (c)}~verification with controlled and uncontrolled images, and
{\bf (d)}~closed-set identification.
We also evaluate the computational cost for LSED generation as well as the query time in closed-set
identification problems.
In addition, 
we use synthetic data to demonstrate the weakness of the holistic SR descriptor (from Section~\ref{sec:src_limitation_verification})
on verification problems.

Experiments were conducted on five datasets:
FERET~\cite{FERET_Protocol},
AR~\cite{Aleix_TR_1998},
BANCA~\cite{BANCA_DATABASE},
exYaleB~\cite{Lee_PAMI_2005},
and ChokePoint~\cite{Wong_CVPRW_2011}.
Figure~\ref{fig:face_images} shows example raw images.
In all experiments,
we used closely cropped face images with a size of $64 \times 64$ pixels.
Each image was manually aligned so that the eyes were at fixed positions,
except for experiments with simulated image variations.
See Figure~\ref{fig:face_crop_example} for examples.

In the following subsections,
we denote the original formulation of MRH with probabilistic encoding as \mbox{LSED+prob}. 
Forms of LSED with the Sparse Autoencoder Neural Network and $l_1$-minimisation based encoding approaches
are denoted as \mbox{LSED+SANN} and \mbox{LSED+$l_1$}, respectively.
The LSED framework has a number of parameters that affect performance.
Based on preliminary experiments,
we split each image into $3 \times 3$ regions and used 32 cohorts for the distance normalisation in Eqn.~(\ref{eqn:norm_dist}). 
LSED+SANN has 512 hidden units, 
where the parameters of the cost function (Eqns.~(\ref{eqn:SANN_cost}) and~(\ref{eqn:SANN_cost_2}))
were set as $\beta = 3$, $\rho = 0.1$, and $\lambda = 0.01$.
\mbox{LSED+prob} and \mbox{LSED+$l_1$} have a dictionary with 1024 Gaussians/atoms.
The threshold for reconstruction error, $\epsilon$, in Eqn.~(\ref{eqn:l1_sparse}) was set to 0.1. 
These parameters were kept unchanged for all experiments.

Unless otherwise specified,
all experiments were implemented in {MATLAB} using an in-house implementation.
The $l_1$-minimisation problem 
was solved with SparseLab\footnote{SparseLab is available at http://sparselab.stanford.edu/}.

\subsection{Face Verification Experiments}
\label{sec:experiments_verification}
 
In each of the following verification experiments,
the face images were divided into three sets:
(1)~training set, 
(2)~development set, and 
(3)~evaluation set.
For all experiments,
except the verification experiment on the BANCA dataset,
we exclusively used the CAS-PEAL dataset~\cite{Gao_TSMC_2008} as the training set.
The CAS-PEAL dataset provides 1200 face images from 1200 unique individuals.
Note that the face images for cohort normalisation are selected from the training set.
The development and evaluation sets have a balanced number of matched and mismatched pairs.

Using the development set and the normalised matching scores from Eqn.~(\ref{eqn:norm_dist}), 
we obtained a decision threshold, $\tau_D$,
which was then used on the evaluation set for assessing the final accuracy.
Specifically, the threshold was adjusted such that
the False Acceptance Rate (FAR) and False Rejection Rate (FRR)
on the development set were equal (\ie, the so-called Equal Error Rate point~\cite{Doddington_SC_2000}).
The threshold was then applied on the evaluation set,
with the final accuracy defined as
{\small $1 - \frac{1}{2}(\operatorname{FAR} + \operatorname{FRR})$}.
The threshold was deliberately not found on the evaluation set
as in real-life conditions it has to be selected {\it a priori}~\cite{Cardinaux_TSP_2006,Bengio_2004}.

In all experiments,
we compared LSED with the holistic SR descriptor described in
Section~\ref{sec:src_limitation_verification}. 
We used the holistic SR descriptor in conjunction with two feature extraction methods:
{\bf (1)}~PCA based~\cite{Belhumeur_PAMI_1997} (denoted as \mbox{PCA+SR}), and
{\bf (2)}~Gabor based~\cite{Liu_TIP_2002} (denoted as \mbox{Gabor+SR}).
Based on preliminary experiments,
the similarity scores between two \mbox{PCA+SR} descriptors were calculated via Hamming distance measurement, 
whereas Euclidean distance was preferred for \mbox{Gabor+SR}. 
Gabor based feature extraction followed the configuration in~\cite{Yang_ECCV_2010}, with PCA based dimensionality reduction.
For both feature extraction methods, PCA preserved 99\% of the total energy.

We also evaluated verification performance of three baseline holistic face descriptors (\ie,~without sparse encoding):
{\bf (1)}~PCA based (denoted~as PCA), 
{\bf (2)}~Local Binary Patterns~\cite{Ahonen_PAMI_2006} (denoted as LBP), and
{\bf (3)}~Gabor based (denoted as Gabor).
The similarities between two face descriptors were calculated using Euclidean distance measurement.

\subsubsection{Face Verification with Alignment Errors and Blurring}
\label{sec:experiments_alignment}

\begin{figure*}[!tb]
  \centering
  \begin{minipage}{0.45\textwidth}
    \centering
    \begin{minipage}{0.15\columnwidth}
      \centerline{\includegraphics[width=\columnwidth]{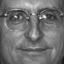}}
      \vspace{-0.8ex}
      \centerline{\footnotesize Aligned}
      \vspace{-0.8ex}
      \centerline{\footnotesize ~}
    \end{minipage}
    \hfill
    \begin{minipage}{0.15\columnwidth}
      \centerline{\includegraphics[width=\columnwidth]{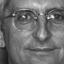}}
      \vspace{-0.8ex}
      \centerline{\footnotesize Horizontal}
      \vspace{-0.8ex}
      \centerline{\footnotesize Shift}
    \end{minipage}
    \hfill
    \begin{minipage}{0.15\columnwidth}
      \centerline{\includegraphics[width=\columnwidth]{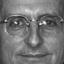}}
      \vspace{-0.8ex}
      \centerline{\footnotesize Vertical}
      \vspace{-0.8ex}
      \centerline{\footnotesize Shift}
    \end{minipage}
    \hfill
    \begin{minipage}{0.15\columnwidth}
      \centerline{\includegraphics[width=\columnwidth]{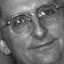}}
      \vspace{-0.8ex}
      \centerline{\footnotesize In-Plane}
      \vspace{-0.8ex}
      \centerline{\footnotesize Rotation}
    \end{minipage}
    \hfill
    \begin{minipage}{0.15\columnwidth}
      \centerline{\includegraphics[width=\columnwidth]{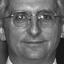}}
      \vspace{-0.8ex}
      \centerline{\footnotesize Scale}
      \vspace{-0.8ex}
      \centerline{\footnotesize Change}
    \end{minipage}
    \hfill
    \begin{minipage}{0.15\columnwidth}
      \centerline{\includegraphics[width=\columnwidth]{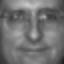}}
      \vspace{-0.8ex}
      \centerline{\footnotesize Blurring}
      \vspace{-0.8ex}
      \centerline{\footnotesize ~}
    \end{minipage}
    
    \caption
      {
      Examples of simulated image variations on FERET.
      }
    \label{fig:feret_align_sample}
  \end{minipage}
  ~
  \begin{minipage}{0.45\textwidth}
    \centering
    \begin{minipage}{0.84\columnwidth}
      \begin{minipage}{0.18\columnwidth}
        \centerline{\includegraphics[width=\columnwidth]{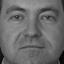}}
        \vspace{-0.8ex}
        \centerline{\footnotesize Frontal}
      \end{minipage}
      \hfill
      \begin{minipage}{0.18\columnwidth}
        \centerline{\includegraphics[width=\columnwidth]{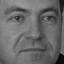}}
        \vspace{-0.8ex}
        \centerline{\footnotesize $+15^\circ$}
      \end{minipage}
      \hfill
      \begin{minipage}{0.18\columnwidth}
        \centerline{\includegraphics[width=\columnwidth]{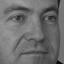}}
        \vspace{-0.8ex}
        \centerline{\footnotesize $+25^\circ$}
      \end{minipage}
      \hfill
      \begin{minipage}{0.18\columnwidth}
        \centerline{\includegraphics[width=\columnwidth]{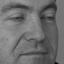}}
        \vspace{-0.8ex}
        \centerline{\footnotesize $+45^\circ$}
      \end{minipage}
      \hfill
      \begin{minipage}{0.18\columnwidth}
        \centerline{\includegraphics[width=\columnwidth]{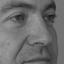}}
        \vspace{-0.8ex}
        \centerline{\footnotesize $+60^\circ$}
      \end{minipage}
    \end{minipage}

    \caption
      {
      Examples of the FERET pose subset.
      }
    \label{fig:feret_pose_sample}
  \end{minipage}
\end{figure*}

In this section, 
we evaluate the robustness of LSED on blurring, 
as well as on four alignment errors
using images taken from the `fb' subset of FERET.
Example images are shown in Figure~\ref{fig:feret_align_sample}.
The generated alignment errors%
\footnote
  {
  The generated alignment errors are representatives
  of real-life characteristics of automatic face localisation/detection algorithms~\cite{Rodriguez_IVC_2006}.
  }
are:
horizontal shift and vertical shift
(using displacements of
{\small $\pm2$}, {\small $\pm4$}, {\small $\pm6$}, {\small $\pm8$} pixels),
in-plane rotation
(using rotations of {\small $\pm10^{\circ}$}, {\small $\pm20^{\circ}$}, {\small $\pm30^{\circ}$}),
and scale variations (using scaling factors of
{\small $0.7$}, {\small $0.8$}, {\small $0.9$}, {\small $1.1$}, {\small $1.2$}, {\small $1.3$)}.
To simulate variations in sharpness,
each original image was first downscaled to three sizes
({\small $48\times48$}, {\small $32\times32$} and {\small $16\times16$} pixels),
and then rescaled to the baseline size of {\small $64\times64$} pixels.
Using the frontal subset `ba' and the expression subset `bj', 
we randomly generated 800 matched and mismatched pairs for each alignment error.
The experiments were conducted with 5-fold validations.
We report the mean accuracy for each scenario.

\begin{figure*}[!tb]
\centering
  \includegraphics[width=2.0\columnwidth]{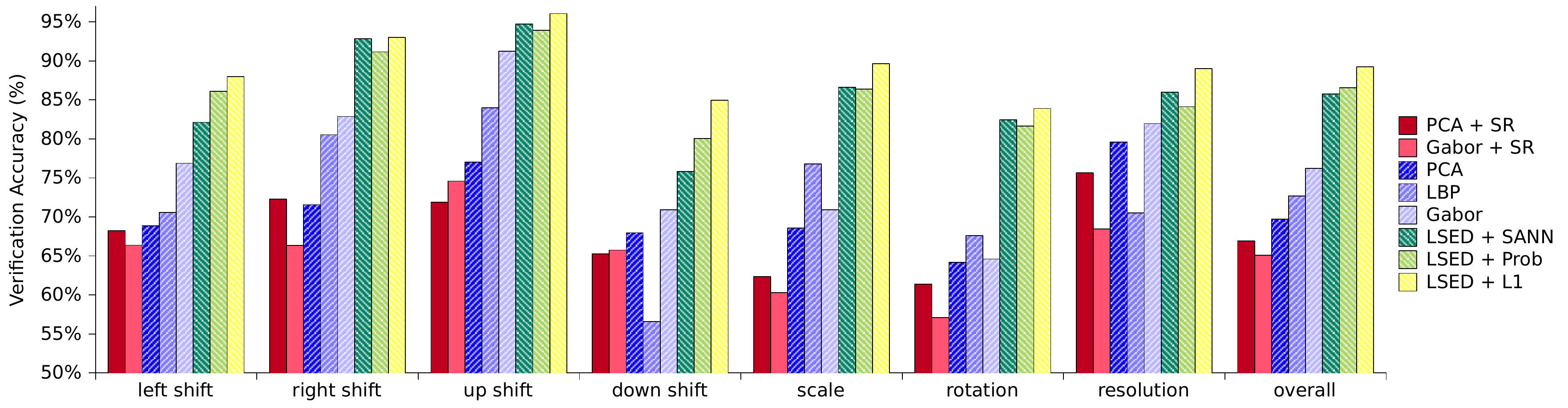}
  \vspace{-2ex}
  \caption
    {
    The average verification accuracy on FERET images with stimulated alignment errors and
    sharpness variations 
    (demonstrated in Fig.~\ref{fig:feret_align_sample}).
    Experiments were conducted with 5-fold validations.
    }
  \label{fig:rec_face_alignment}
\end{figure*}

The results,
presented in Figure~\ref{fig:rec_face_alignment},
show that the three LSED approaches consistently achieved robust performance in all simulated scenarios.
\mbox{LSED+SANN} and \mbox{LSED+prob} achieved average accuracies of 85.8\% and 86.2\%,
respectively,
whereas \mbox{LSED+$l_1$} led the performance with an average accuracy of 89.2\%.
Overall, 
the accuracy of \mbox{LSED+$l_1$} is about 12.2 percentage points better than the baseline Gabor approach
and about 23.7 percentage points when compared to \mbox{Gabor+SR}.
The results also show that \mbox{PCA+SR} and \mbox{Gabor+SR} performed poorly on all misalignment errors, 
with overall accuracies of 68.2\% and 65.6\%, respectively.
The results suggest that scale changes and in-plane rotation variations are in general the hardest problems out of all alignment errors.

\begin{figure*}[!tb]
\centering
  \includegraphics[width=2.0\columnwidth]{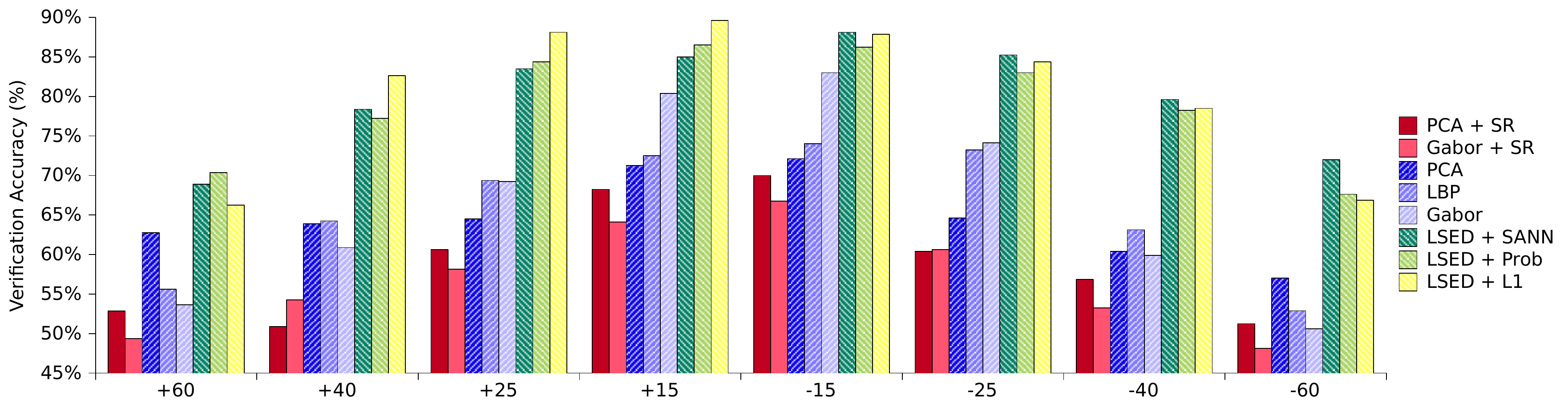}
  \vspace{-2ex}
  \caption
    {
    Verification performance on pose mismatches for various angles.
    Faces from each pose angle are compared with the FERET frontal subset `ba' and the expression subset `bj'.
    Experiments were conducted with 5-fold validations.
    }
  \label{fig:rec_face_pose}
\end{figure*}

\subsubsection{Face Verification with Pose Mismatches}
\label{sec:experiments_pose_alignment}

In this section we evaluate the robustness of LSED for handling pose mismatches.
We selected the `b' subset from the FERET dataset,
which has 200 images per pose.
The evaluation process on each pose angle was the same as the method described in the previous section.
Example images are shown in Figure~\ref{fig:feret_pose_sample}.

The results,
shown in Figure~\ref{fig:rec_face_pose},
indicate that the three LSED approaches considerably outperforms both \mbox{PCA+SR} and \mbox{Gabor+SR}.
Both of the holistic SR descriptors obtained a maximum accuracy of 56.9\%
when the absolute value of the pose angle was {\small $\geq 40^{\circ}$}. 
In contrast,
\mbox{LSED+$l_1$} achieved an average accuracy of 73.6\% under the same pose angles.
Note that the \mbox{LSED+$l_1$} was outperformed by \mbox{LSED+SANN} and \mbox{LSED+prob} for pose angles of {\small $\pm60^{\circ}$}. 
When the pose angle was between {\small $-25^{\circ}$} and {\small $+25^{\circ}$} 
(\ie~relatively frontal) 
the best performing holistic SR descriptor (\mbox{PCA+SR}) achieved an average accuracy of about 64.8\%.
All LSED approaches outperformed the holistic SR descriptors by a comfortable margin on the same range of pose angles, 
with \mbox{LSED+$l_1$} obtaining an average accuracy of 87.5\%.

\subsubsection{Face Verification with Frontal Faces}
\label{sec:experiments_verification_1}

In this experiment,
we evaluated the performance on three datasets with images captured in various environment
conditions.
Example images are shown in Figure~\ref{fig:face_crop_example}.
The first dataset is AR~\cite{Aleix_TR_1998},
which contains 100 unique subjects with 14 images per subject.
We randomly generated 9800 pairs of matched and mismatched pairs and evaluated the performance of
each algorithm with 5-fold validations.
The second dataset is BANCA~\cite{BANCA_DATABASE}.
We report only the results on the `P' protocol,
where the algorithm was trained in controlled conditions and tested on a combination of controlled,
degraded and adverse images.
According to the protocol,
the 52 subjects were divided into two groups,
where each group played the role of the development set and
evaluation set in turn.
We randomly selected one image per person from each video.
The third dataset is ChokePoint~\cite{Wong_CVPRW_2011},
which was recorded under real-world surveillance conditions.
It has 16 videos of 29 subjects recorded on four distinct portals%
\footnote
  {
  A portal is a location where a camera rig is placed to capture faces from multiple
  angles. 
  Each portal has a unique background and lighting conditions.
  }.
We randomly generated 38,710 matched and mismatched image pairs where each pair consisted of images
taken from different portals (\ie~cross environment matching).
The experiments were evaluated with 5-fold validations.

\begin{table}[!t]
  \centering
  \caption
    {
    Frontal face verification performance on several datasets.
    The face images were closely cropped to exclude hair and background, 
    and scaled to $64 \times 64$ pixels. 
    The values in {\bf bold} indicate the best performing algorithm for each dataset.
    }
  \label{tab:face_ver_experiment}
  \begin{tabular}{l | c c c | c}
  \toprule
  {\bf Method}        & {\bf ~ AR ~ }     
                      & {\bf BANCA } 
                      & {\bf ChokePoint }  
                      & {\bf ~ Overall} \\
  \midrule
  PCA + SR            & 61.4\%      & 58.8\%      & 57.4\%      & 59.4\% \\ 
  Gabor + SR          & 66.1\%      & 63.3\%      & 59.5\%      & 63.2\% \\ 
  \midrule
  PCA                 & 57.3\%      & 63.5\%      & 55.6\%      & 59.0\% \\ 
  LBP                 & 77.9\%      & 60.3\%      & 65.3\%      & 68.1\% \\ 
  Gabor               & 74.5\%      & 70.0\%      & 75.6\%      & 73.2\% \\ 
  \midrule 
  LSED + SANN         & 72.2\%      & 73.4\%      & 75.1\%      & 73.5\% \\ 
  LSED + prob         & 76.8\%      & 75.4\%      & 76.8\%      & 76.3\% \\ 
  LSED + $l_1$        & {\bf 80.0\%}&{\bf 82.0\%} &{\bf 79.8\%} & {\bf 80.7\%} \\ 
  \bottomrule
  \end{tabular}
\end{table}

The results,
presented in Table~\ref{tab:face_ver_experiment},
show that the three LSED methods obtained the best overall performance.
Both \mbox{PCA+SR} and \mbox{Gabor+SR} performed at their best on the laboratory captured AR dataset
and considerably poorer on the more realistic ChokePoint dataset.
The results also show that both the baseline LBP and Gabor methods outperformed the holistic SR descriptors.
For example, the baseline Gabor approach obtained an overall accuracy of 73.2\%,
outperforming its sparse counterpart (Gabor+SR) which obtained an overall accuracy of 63.2\%.

\mbox{LSED+$l_1$} achieved the best overall accuracy of 80.7\%.
On the controlled AR dataset, 
The baseline LBP method outperformed both \mbox{LSED+SANN} and \mbox{LSED+prob} by 5.7 and 1.1 percentage points, respectively.
However,  the performance of LBP dropped considerably on both the BANCA and ChokePoint datasets,
where \mbox{LSED+prob} outperformed LBP by 15.1 and 11.5 percentage points on the corresponding datasets.
This indicates that while the LSED framework can be outperformed by baseline holistic methods in controlled conditions,
LSED is more robust for face images obtained in uncontrolled conditions.

\subsubsection{Experiments with Synthetic Data}
\label{sec:experiments_synthetic_data}

The results obtained in the preceding sections indicate that holistic SR descriptors
were consistently outperformed by baseline holistic face descriptors (\ie, without sparse coding).
In this section, 
we performed a set of verification experiments with synthetic data to study this phenomenon further.

We explicitly created a dictionary {\small $\Mat{D}$} which does not satisfy the underlying {\it sparsity} assumption.
Each sample from the synthetic data is assumed to be a holistic representation of a face.
The synthetic data comprised of 232 random classes,
with the samples in each class obeying a normal distribution. 
The dimensionality of data was 16. 
For each class 128 samples were generated.
We randomly selected 32 classes as the training set and the remaining 200 classes as the development
set and evaluation set.
The training set played the role of dictionary {\small $\Mat{D}$} in Eqn.~(\ref{eqn:sr_verification}).
The experiments were conducted with 5-fold validations.

Several verification experiments with increasing difficulty were generated
by fixing the mean of each class and increasing the class variance.
The distribution of the class means was carefully controlled such that at the smallest class variance
the mutual overlaps between classes are close to zero.
We employed direct feature matching as the baseline.
In other words, 
for two given samples, {\small $\Vec{x}_a$} and {\small $\Vec{x}_b$},
the matching score is the Euclidean distance \mbox{\small $\|\Vec{x}_a-\Vec{x}_b\|_2$}.
The holistic SR descriptor was evaluated with Hamming distance measurement,
as this led to somewhat better performance than using the Elucidean distance.
The Hamming distance compares two descriptors by measuring if the corresponding descriptors have the same set of nonzero entries.
In other words,
Hamming distance explicitly inspects if both descriptors are spanned by a set of common subspaces.
 
\begin{figure}[!tb]
\centering
  \includegraphics[width=1.0\columnwidth]{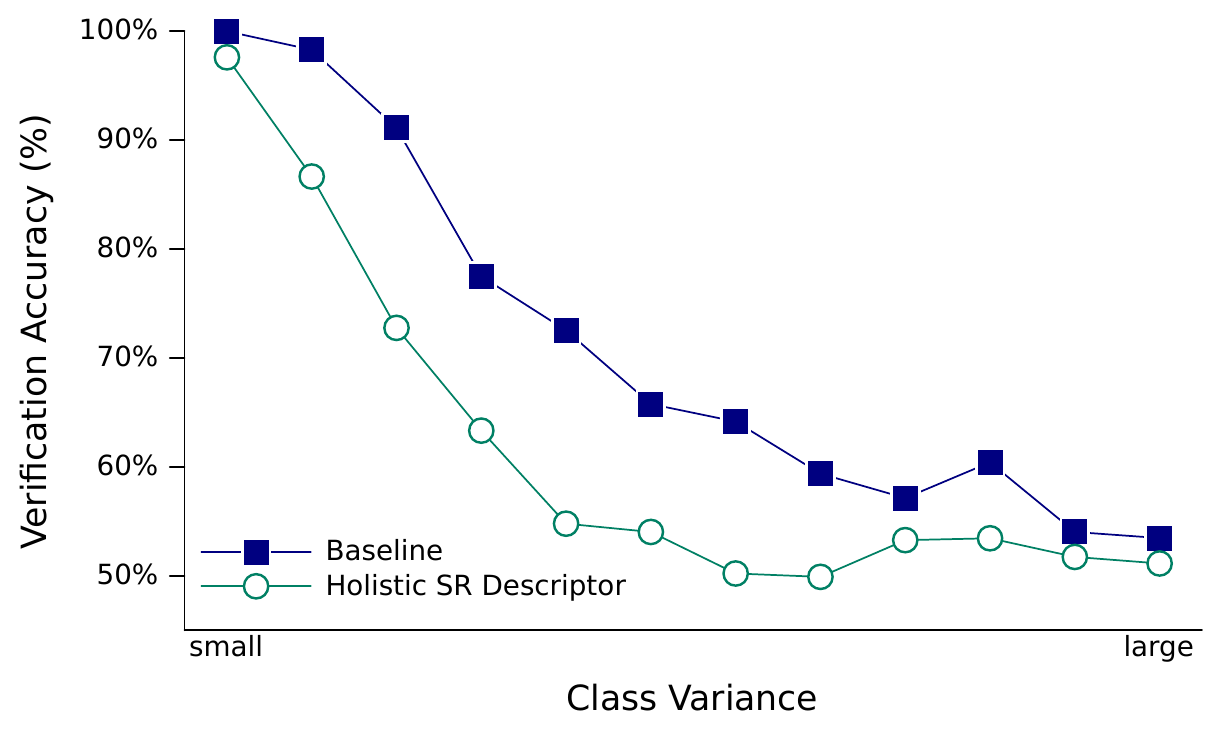}
  \caption
    {
    \small
    Verification performance on synthetic data.
    Experiments were conducted with varying class variance,
    where large class variance indicates strong overlap between classes.
    The baseline was achieved by matching each feature pair using Euclidean distance.
    Experiments were conducted with 5-fold validations.
    }
  \label{fig:synthetic_experiment}
\end{figure}

The results in Figure~\ref{fig:synthetic_experiment} show that the baseline performance is close to 100\% when the class variance is small, 
and drops to 53.5\% when variance is at its maximum value.
In contrast, 
the holistic SR descriptor achieved poorer performance across the variance range,
with accuracies of 97.6\% and 51.1\% respectively for minimum and maximum class variance.
This result agrees with our discussion in Section~\ref{sec:src_limitation}
and the findings from the preceding face verification experiments.
Specifically, if the class information of the atoms is not given
and the sparsity assumption does not hold for the dictionary {\small $\Mat{D}$}, 
the resulting sparse solutions do not provide good discriminative ability when compared to the original holistic representation.

\subsection{Face Identification Experiments}
\label{sec:experiments_identification}

In the preceding set of experiments,
we demonstrated that the proposed LSED framework outperforms holistic SR descriptors on various face verification problems.
In this section,
we evaluate the efficacy of LSED in closed-set face identification,
which is the identity inference configuration typically used in SR related literature.

LSED was compared with five established holistic SR based classification algorithms:
{\bf   (i)}~SR with PCA feature extraction (denoted as PCA+SRC)~\cite{Wright2009_PAMI},
{\bf  (ii)}~SR with PCA feature extraction and LDA (denoted as LDA+SRC)~\cite{Wright2009_PAMI},
{\bf (iii)}~SR with Gabor feature extraction (denoted as Gabor+SRC)~\cite{Yang_ECCV_2010},
{\bf  (iv)}~Robust Sparse Coding with PCA feature extraction (denoted as RSC)~\cite{Yang_CVPR_2011},
and
{\bf   (v)}~orthonormal $l_2$-norm approach with vectorised raw image~\cite{Shi_CVPR_2011} (denoted as raw+$l_2$).
Instead of solving an optimisation problem,
\mbox{raw+$l_2$} estimates the sparse code {\small $\Vec{\alpha}$} using
\mbox{\small $\Vec{\alpha} = \Mat{R}^{-1}\Mat{Q}^{T}\Vec{x}$},
where {\small $\Mat{Q}$} and {\small $\Mat{R}$} are the result of QR factorisation~\cite{Trefethen_1997} of dictionary {\small $\Mat{D}$}.

The experiments were conducted on AR, exYaleB and ChokePoint datasets,
with each gallery having 7, 16, and 16 images per class, respectively.
To increase the difficulty,
the gallery of the ChokePoint dataset was selected from a portal different than the portal used for the query images.
Each portal has a unique background and illumination conditions.
The identification performance of LSED was obtained with the Nearest Neighbour classifier.
Note that the results shown for the established SR algorithms are slightly different from the literature,
due to the image size and dataset splits being different.

\begin{table*}[!t]
  \centering
  \caption
    {
    Closed-set identification performance of the proposed method and various SR based approaches.
    The values in brackets are the number of images per class in the gallery.
    The values in {\bf bold} indicate the best performing algorithm for each dataset.
    }
  \label{tab:face_rec_experiment}
  \begin{small}
  \begin{tabular}{l | c c c | c}
  \toprule
  {\bf Method}  & {\bf ~ ~ AR (7) ~ }
                & {\bf exYaleB (16)}
                & {\bf ChokePoint (16)}
                & {\bf ~ Overall ~}  \\
  \midrule
  PCA + SRC~\cite{Wright2009_PAMI}  & 81.0\%      & 67.9\%      & 17.5\%     & 52.1\% \\ 
  LDA + SRC~\cite{Wright2009_PAMI}  & 89.7\%      & 52.8\%      & 65.3\%     & 64.9\% \\ 
  Gabor + SRC~\cite{Yang_ECCV_2010} & 91.7\%      & 61.4\%      & 63.4\%     & 68.3\% \\ 
  RSC~\cite{Yang_CVPR_2011}         & 95.7\%      & 72.8\%      & 64.5\%     & 74.4\% \\ 
  raw + $l_2$~\cite{Shi_CVPR_2011}  
                                    & 90.3\%      & 75.1\%      & 76.0\%     & 78.5\% \\ 
  \midrule
  LSED + SANN                       & 96.3\%      & 66.0\%      & 77.0\%     & 76.2\% \\ 
  LSED + prob                       & 97.9\%      & 76.7\%      & 80.5\%     & 82.4\% \\ 
  LSED + $l_1$                      & {\bf 98.9\%}& {\bf 90.9\%}&{\bf 85.0\%}&{\bf 90.4\%} \\ 
  \bottomrule
  \end{tabular}
  \end{small}
\end{table*}

The results,
shown in Table~\ref{tab:face_rec_experiment},
indicate that \mbox{LSED+prob} and \mbox{LSED+$l_1$} consistently outperformed all SRC algorithms in closed-set identification.
The improvement on the ChokePoint dataset is the most notable among the three datasets,
where \mbox{LSED+$l_1$} outperformed the closest SRC algorithm (\ie~RSC) by 20.5 percentage points.
It also outperformed the \mbox{raw+$l_2$} approach by 9 percentage points.

\subsection{Computation Time}

The preceding verification and identification experiments indicate
that the \mbox{LSED+$l_1$} technique achieves the best overall performance.
However, the superior performance of \mbox{LSED+$l_1$} comes at the expense of considerably higher computational cost.
As shown in Table~\ref{tab:LSED_time},
\mbox{LSED+$l_1$} requires 7739 milliseconds (ms) to generate a single face descriptor,
mainly due to solving multiple expensive \mbox{$l_1$-minimisation} problems (one for each small patch).
In contrast, 
\mbox{LSED+SANN} is approximately 70 times faster, as it requires only 2.9ms to generate the entire face descriptor.
\mbox{LSED+prob}, which achieved the closest performance to \mbox{LSED+$l_1$},
requires approximately a quarter of time when compared with \mbox{LSED+$l_1$}.
We note that the computation cost for all three LSED variants can be considerably reduced via parallelisation,
as each patch can be processed independently prior to the pooling operation in Eqn.~(\ref{eqn:avg_pooling}).

\begin{table}[!t]
  \centering
  \caption
    {
    Average computation time for generating a Locally Sparse Encoded Descriptor for one image.
    }
  \label{tab:LSED_time}
  \begin{small}
  \begin{tabular}{l c}
  \toprule
  {\bf Method} & {\bf Time (milliseconds)} \\
  \midrule
  LSED + SANN  & ~ 110 \\
  LSED + prob  &  2021 \\
  LSED + $l_1$ &  7739 \\ 
  \bottomrule
  \end{tabular}
  \end{small}
\end{table}

\begin{figure}[!tb]
\centering
  \includegraphics[width=0.9\columnwidth]{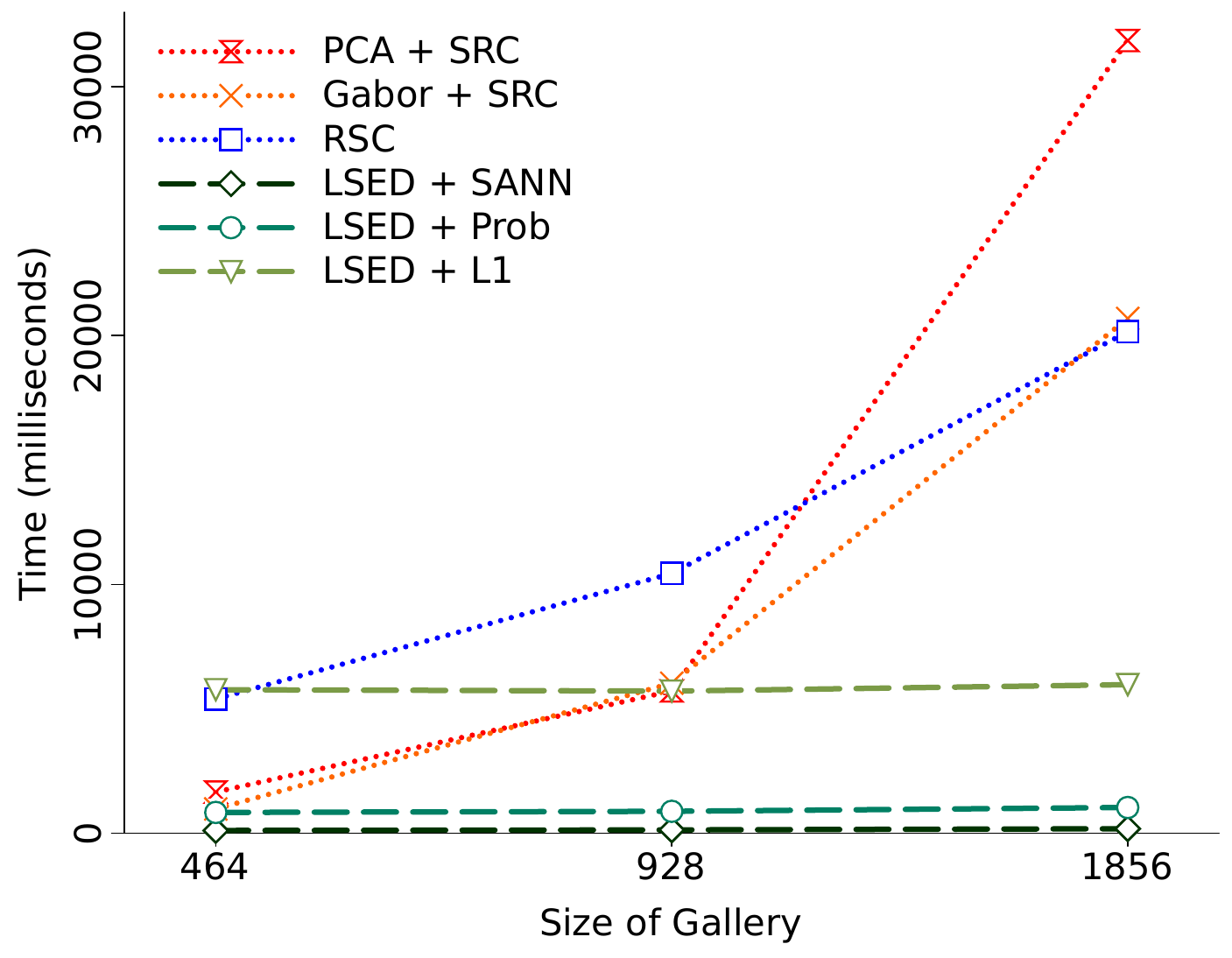}
  \caption
    {
    \small
    Average computation time
    (including feature extraction, sparse encoding and identification)
    for matching a probe image against galleries of various sizes.
    }
  \label{fig:face_rec_experiment_time}
\end{figure}

Other than the computational cost generating each face descriptor,
the cost to match one probe against a large gallery is also important.
Using galleries with various amount of face images,
we evaluated the average time to recognise a single probe using a closed-set identification setup.
For each method,
we measured the time for feature extraction, sparse encoding (or approximate sparse solution) and identification.
The \mbox{raw+$l_2$} method is not included in this evaluation as it does not solve an optimisation problem. 

The results,
shown in Figure~\ref{fig:face_rec_experiment_time},
indicate that the identification time for LSED framework is almost constant.
In contrast, 
the computational cost of traditional SRC-based methods and RSC increased considerably as the gallery size increased.

\section{Experiments with Image Sets}
\label{sec:experiments_image_set}

In the previous section we presented experiments using a single face image per person at a time.
In contrast, in this section we evaluate the verification performance of LSED using multiple images per person at a time.
This recognition task is also known as image set matching,
with the aim of determining if two face sets, 
$\mathbb{A}$ and $\mathbb{B}$,
belong to the same person.

We first describe two image set matching approaches
(Hausdorff distance and mean descriptors),
followed by presenting results on BANCA and ChokePoint datasets.
We also contrast the computational costs of the two matching approaches.

\subsection{Image Set Matching via Hausdorff Distance}
\label{sec:experiments_image_set_hausdorff}

Given two finite image sets, 
$\mathbb{A} = \{ a_1, a_2, \ldots, a_{N_\mathbb{A}}\}$
and
$\mathbb{B} = \{ b_1, b_2, \ldots, b_{N_\mathbb{B}}\}$,
the Hausdorff distance is defined as:
\begin{equation}
  H(\mathbb{A},\mathbb{B}) 
  = 
  \operatorname{max} 
    \: \{ \:  
    h(\mathbb{A},\mathbb{B}), 
    h(\mathbb{B},\mathbb{A}) 
    \: \}
\label{eqn:hausdorff_1}
\end{equation}%
where
\begin{equation}
  h(\mathbb{A},\mathbb{B}) 
  = 
  \underset{i \in \mathbb{A}}{\operatorname{max}} 
    \: \{ \:  
    \underset{j \in \mathbb{B}}{\operatorname{min}} 
      \:  \{ 
      s(a_i,b_j)
      \}
    \: \}
\label{eqn:hausdorff_2}
\end{equation}

\noindent
and $s(\cdot)$ measures the similarity between two images. 
The function $h(\mathbb{A},\mathbb{B})$ is called the directed Hausdorff distance from $\mathbb{A}$ to $\mathbb{B}$.
In general,
if the Hausdorff distance between image set $\mathbb{A}$ and $\mathbb{B}$ is $d$,
each image in $\mathbb{A}$ is within distance $d$ to some of the points in $\mathbb{B}$, 
and vice-versa~\cite{Huttenlocher_PAMI_1993}.

\subsection{Image Set Matching with Mean Descriptors}
\label{sec:experiments_image_set_mean}

The Hausdorff distance measurement is a computationally expensive approach for image set matching.
This is in particularly a problem for video surveillance of public spaces,
where the volume of surveillance video can be very high.
To address this problem,
each image set can be represented by an overall descriptor via straightforward averaging of the corresponding face descriptors~\cite{Kang_EURASIP_2010}.
Specifically, given descriptors from image set~{\small $\mathbb{A}$}, 
the mean descriptor is represented as
$\frac{1}{N_\mathbb{A}} \sum\nolimits_{n=1}^{N_\mathbb{A}} \Vec{h}_{\mathbb{A},n}$,
where $\Vec{h}_{\mathbb{A},n}$ is the $n$-th descriptor of~$\mathbb{A}$.
The similarity between two mean descriptors can be then computed using Eqn.~(\ref{eqn:norm_dist}).

In contrast to image set matching using the Hausdorff distance,
the total number comparisons between $\mathbb{A}$ and $\mathbb{B}$
is reduced from  $N_\mathbb{A} \times N_\mathbb{B}$ to one.

\subsection{Results}
\label{sec:experiments_image_set_results}

\begin{figure}
  \begin{center}
  \begin{minipage}{1.0\columnwidth}
    \begin{minipage}{1.0\columnwidth}
      \parbox[b][26ex][c]{2ex}{\small\bf (a)}
      \includegraphics[width=0.99\columnwidth]{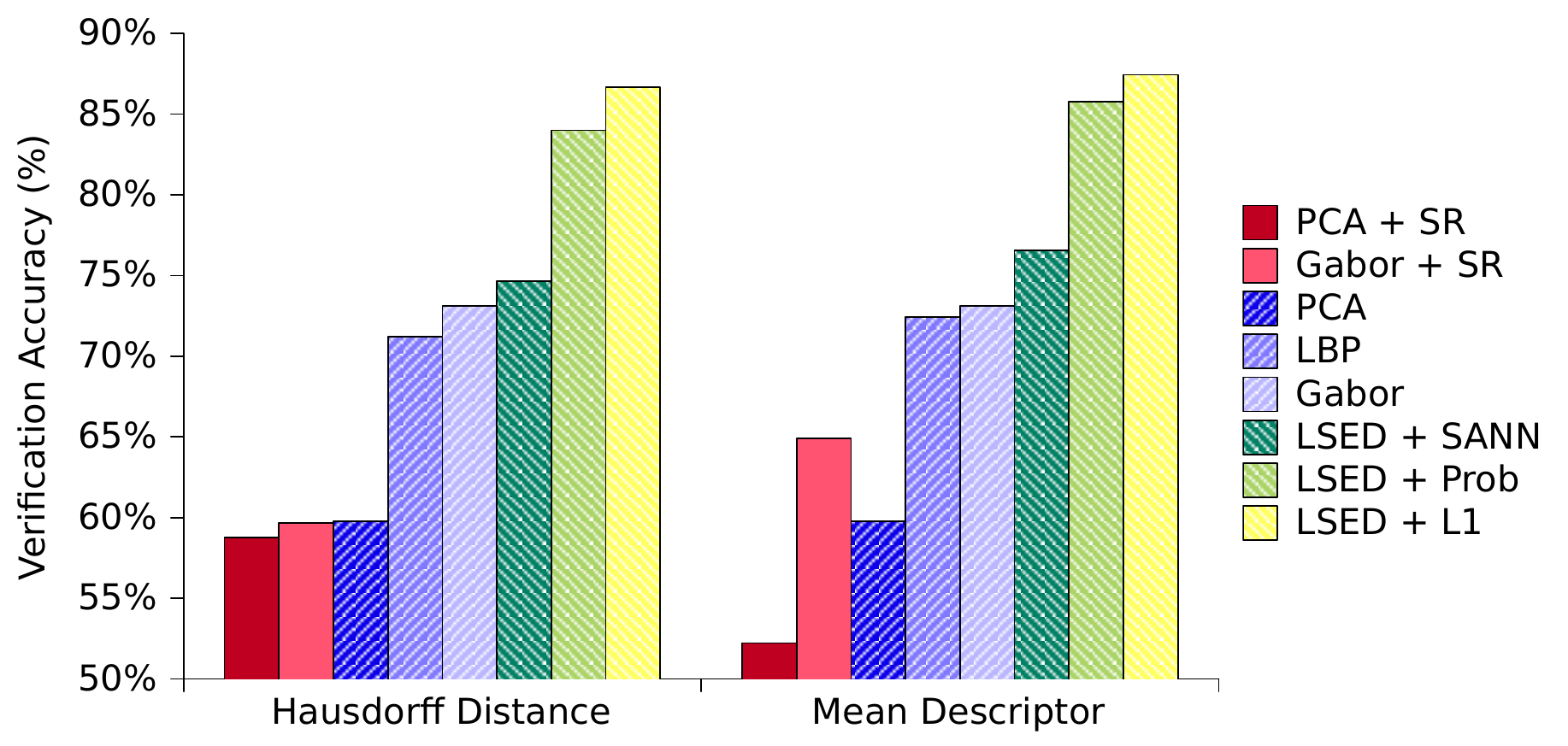}
    \end{minipage}
    \begin{minipage}{1.0\columnwidth}
      \parbox[b][26ex][c]{2ex}{\small\bf (b)}
      \includegraphics[width=0.99\columnwidth]{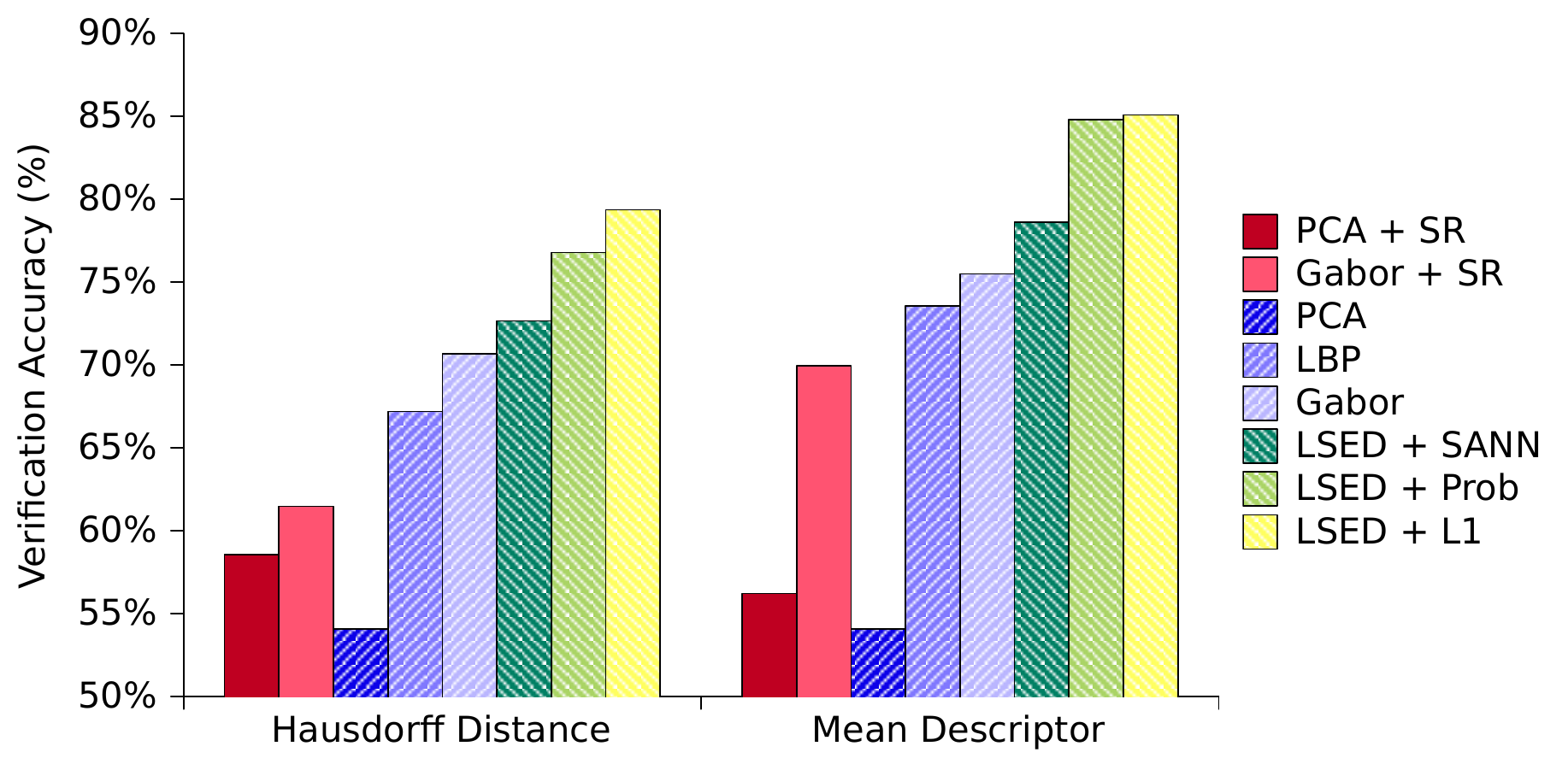}
    \end{minipage}
  \end{minipage}
  \end{center}
  \caption
    {
    Image set verification performance on  
    {\bf (a)} BANCA dataset with 5-fold validations, and
    {\bf (b)} ChokePoint dataset with 10-fold validations.
    }
  \label{fig:rec_image_set}
\end{figure}

We evaluate image set matching performance on two datasets,
with images captured under uncontrolled environment conditions.
The first dataset is BANCA dataset,
where we randomly generate 900 pairs of matched and mismatched pairs, and each image-set contains 9 face images.
The experiments were evaluated with 5-fold validations.
The second dataset is the ChokePoint video dataset.
We selected 16 images with the highest quality as per~\cite{Wong_CVPRW_2011}
and randomly generated 5000 matched and mismatched pairs.
The experiments were evaluated using 10-fold validations.
For comparison, we used the same face descriptor methods as in Section~\ref{sec:experiments_verification}.
The results are shown in Figure~\ref{fig:rec_image_set}.

On the BANCA dataset, 
the performance of the three LSED approaches is very similar for both the Hausdorff and mean descriptor matching approaches.
Among the LSED variants, \mbox{LSED+$l_1$} in conjunction with mean descriptor matching obtains the highest accuracy,
with the computationally less expensive \mbox{LSED+prob} variant not far behind.
The performance of baseline LBP and Gabor approaches are considerably lower than \mbox{LSED+$l_1$}.
\mbox{PCA+SR} and \mbox{Gabor+SR} achieved poor performance for both the Hausdorff and mean descriptor matching approaches.

The verification performance on the ChokePoint dataset has two notable differences
when compared to the performance on the BANCA dataset. 
All LSED variants achieved notably better performance using the mean descriptor matching approach
rather than the Hausdorff distance based approach.
Secondly,
the traditional PCA approach obtained the worst verification accuracy among all face descriptors,
with \mbox{PCA+SR} outperforming it by 4.5 percentage points when using the Hausdorff distance.
The poor performance is mainly due to image quality variations (\ie~stemming from surveillance environments),
which was also shown in the face identification experiments in Section~\ref{sec:experiments_identification}.

The approximate computational cost for the mean and Hausdorff matching approaches,
using \mbox{LSED+$l_1$} descriptors, is shown in Table~\ref{tab:face_rec_image_set_experiment_time}.
The straightforward mean descriptor approach is approximately 2 orders of magnitude faster
than the computationally intensive Hausdorff approach,
while obtaining similar or better results.

\begin{table}[!t]
  \centering
  \caption
    {
    Average time for matching two image sets.
    Each image set contains 32 images.
    The experiments were conducted with \mbox{LSED + $l_1$},
    where the dimensionality of each descriptor is 9216. 
    }
  \label{tab:face_rec_image_set_experiment_time}
  \vspace{-1ex}
  \begin{small}
    \begin{tabular}{l| c}
    \toprule
    {\bf Method}                  & {\bf Time (milliseconds)}  \\
    \midrule
    Matching via Hausdorff distance & ~ 1018 \\
    Matching via mean descriptor    & ~ ~ ~ 7 \\
    \bottomrule
    \end{tabular}
  \end{small}
\end{table}
\section{Main Findings}
\label{sec:conclusions}

Most of the literature on Sparse Representation (SR) for face recognition
has focused on holistic face descriptors in closed-set identification applications.
The underlying assumption in SR-based methods is that each class in the gallery
has sufficient samples and the query lies on the subspace spanned by the gallery of the same class.
Unfortunately, 
such assumption is easily violated in the more challenging face verification scenario,
where an algorithm is required to determine if two faces (where one or both have not been seen before) belong to the same person.

We first discussed why previous attempts with SR might not be applicable to verification problems.
We then proposed an alternative approach to face verification via SR.
Specifically,
we proposed to use explicit SR encoding on local image patches rather than the entire face.
The obtained sparse signals are pooled via averaging to form multiple region descriptors,
which are then concatenated to form an overall face descriptor.
Due to the deliberate loss spatial relations within each region (caused by averaging),
the resulting descriptor is robust to misalignment and various image deformations.
Within the proposed framework,
we evaluated several SR encoding techniques:
\mbox{$l_1$-minimisation}, 
Sparse Autoencoder Neural Network (SANN),
and an implicit probabilistic technique based on Gaussian Mixture Models.

Thorough experiments on AR, FERET, exYaleB, BANCA and ChokePoint datasets
show that the proposed local SR approach obtains considerably better
and more robust performance than several previous state-of-the-art holistic SR methods,
in both verification and closed-set identification problems.
The proposed approach is particularly suited to dealing with face images obtained in difficult conditions,
such as surveillance environments.
The experiments also show that 
\mbox{$l_1$-minimisation} based encoding has a considerably higher computational cost
when compared to SANN-based and probabilistic encoding,
but leads to higher recognition rates.

\section{Acknowledgements}
\label{sec:Acknowledgements} 

NICTA is funded by the Australian Government as represented by the Department of Broadband, Communications and the Digital Economy,
as well as the Australian Research Council through the ICT Centre of Excellence program.

\bibliographystyle{ieee}
\bibliography{references}

\end{document}